\documentclass[lettersize,journal]{IEEEtran}
\usepackage{amsmath,amsfonts}
\usepackage{algorithmic}
\usepackage{algorithm}
\usepackage{array}
\usepackage[caption=false,font=normalsize,labelfont=sf,textfont=sf]{subfig}
\usepackage{textcomp}
\usepackage{stfloats}
\usepackage{url}
\usepackage{verbatim}
\usepackage{graphicx}
\usepackage{cite}
\usepackage{amssymb}
\usepackage{booktabs}
\usepackage{xspace}
\usepackage{enumitem}
\usepackage{float}
\usepackage{bbding}
\usepackage[pagebackref=true,breaklinks=true,colorlinks,bookmarks=false]{hyperref}
\usepackage{soul}
\usepackage{color, xcolor} 
\newcommand{\shortname}{DiffStyler\xspace}

\newcommand{\textprompt}{\mathcal{T}}
\newcommand{\etal}{{\sl et al.}}

\begin{document}

\title{DiffStyler: Controllable Dual Diffusion\\ for Text-Driven Image Stylization}

\author{
Nisha~Huang, Yuxin~Zhang, Fan~Tang, Chongyang~Ma, Haibin~Huang, Weiming~Dong,~\IEEEmembership{Member,~IEEE}, Changsheng~Xu,~\IEEEmembership{Fellow,~IEEE}  
\IEEEcompsocitemizethanks{
\IEEEcompsocthanksitem This work was supported in part by the National Science Foundation of China U20B2070, 61832016, and 62102162, and in part by Beijing Natural Science Foundation under no. L221013. (Corresponding author: Weiming Dong.)
\IEEEcompsocthanksitem N. Huang, Y. Zhang, W. Dong and C. Xu are with MAIS, Institute of Automation, Chinese Academy of Sciences, Beijing 100190, China, and also with School of Artificial Intelligence, University of Chinese Academy of Sciences, Beijing 100049, China. E-mail:\{huangnisha2021,zhangyuxin2020,weiming.dong\}@ia.ac.cn and csxu@nlpr.ia.ac.cn.
\IEEEcompsocthanksitem F. Tang is with Institute of Computing Technology, Chinese Academy of Sciences, Beijing 100190, China, E-mail: tangfan@ict.ac.cn.
\IEEEcompsocthanksitem C. Ma and H. Huang are with Kuaishou Technology, Beijing 100085, China. E-mail:chongyangm@gmail.com, jackiehuanghaibin@gmail.com.
}
}

\markboth{IEEE TRANSACTIONS ON NEURAL NETWORKS AND LEARNING SYSTEMS~2023}%
{Huang \MakeLowercase{\textit{et al.}}: DiffStyler: Controllable Dual Diffusion for Text-Driven Image Stylization}


\maketitle

\begin{abstract}
Despite the impressive results of arbitrary image-guided style transfer methods, text-driven image stylization has recently been proposed for transferring a natural image into a stylized one according to textual descriptions of the target style provided by the user.
Unlike the previous image-to-image transfer approaches, text-guided stylization progress provides users with a more precise and intuitive way to express the desired style. 
However, the huge discrepancy between cross-modal inputs/outputs makes it challenging to conduct text-driven image stylization in a typical feed-forward CNN pipeline. 
In this paper, we present DiffStyler, a dual diffusion processing architecture to control the balance between the content and style of the diffused results. 
The cross-modal style information can be easily integrated as guidance during the diffusion process step-by-step. 
Furthermore, we propose a content image-based learnable noise on which the reverse denoising process is based, enabling the stylization results to better preserve the structure information of the content image.
We validate the proposed DiffStyler beyond the baseline methods through extensive qualitative and quantitative experiments.  
Code is available at \url{https://github.com/haha-lisa/Diffstyler}.
\end{abstract}

\begin{IEEEkeywords}
Arbitrary image stylization, diffusion, textual guidance, neural network applications.
\end{IEEEkeywords}




\section{Introduction}
\label{sec:intro}

\IEEEPARstart{I}{mage} stylization is appealing in practice as it allows amateurs to turn real-world photos into renderings that mimic the style of artwork without requiring any professional skills.
Based on the extracted style textures, the style transfer methods~\cite{gatyscnnstyle,huangadain,tnnls_style1,tnnls_style2,tnnls_style3,tnnls_style4} migrate the semantic textures of the style images to the content images and generate vivid artwork.
They are followed and improved by later works~\cite{StyleGAN,wang2020collaborative} that have shown substantial value in creative visual design.

While the above image stylization methods are capable of delivering attractive and stable results for artwork, they all require users to provide a style image as a reference for stylization, which leads to the following problems. First, providing an appropriate style image itself increases the complexity and redundancy of the user operation. In addition, style images themselves have strong limitations for color and texture. As a result, the user's need for stylization and artistic creation cannot be adequately expressed through a single image either.
In contrast, expressing the user's artistic needs and aesthetic preferences through text is more in line with the original intention of artistic creation. 
For illustration, the text conveys arbitrary artists, styles, and artistic movements.
This more natural and intuitive way of textual guidance can create more imaginative and unrestricted digital artworks.

\begin{figure}[t]
    \centering
    \includegraphics[width=0.5\textwidth]{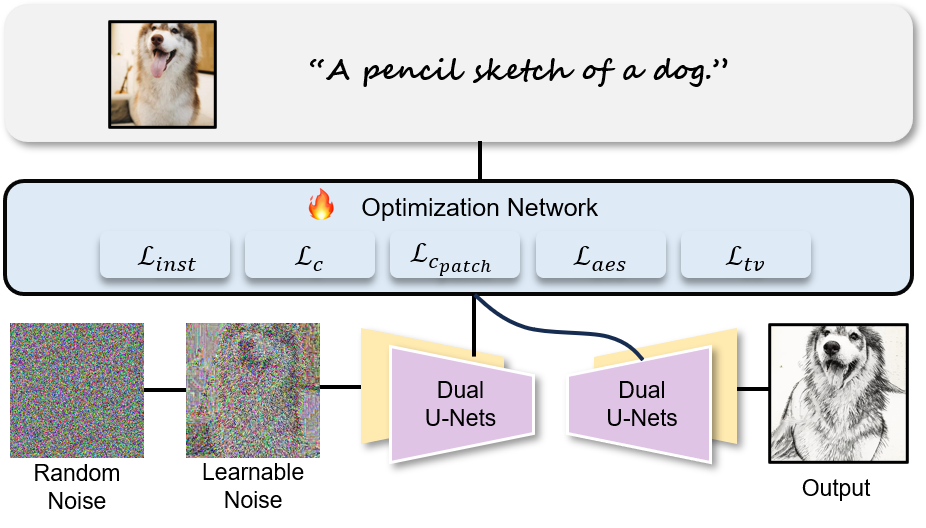}
    \caption{An illustration of a DiffStyler is showcased, wherein the model generates a stylized image by leveraging a text-based description style and the content image as input.}
    \label{fig:schema}
\end{figure}
Current approaches~\cite{StyleCLIP,StyleGANNADA,CLIPstyler} for text-driven stylization tasks are mainly based on GAN~\cite{gan} models, which results in limited modeling and generation capabilities between text and style.
The GAN-based approaches, which rely on the patch style discriminator, are unable to generate satisfactory results in the case of inappropriate patches to sample.
In our investigation, we found that textual information can easily guide the diffusion model for stable stepwise diffusion without relying on additional random style patches.
Meanwhile, with the development of diffusion-based methods~\cite{Guided-Diffusion,dalle2,nichol2021glide,latentdiffusion}, 
they have demonstrated phenomenal results on visual tasks, especially in generating artworks~\cite{Huang2022MGAD,wu2022creative,gal2022image}. 
Therefore, we introduce diffusion into stylization tasks to guide the image global sampling process.
This improves the problem that GAN-based methods~\cite{fu2022ldast,CLIPstyler} repeatedly have the same stylized patches at different locations of the generated results thus causing artifacts and failure to highlight the main content.

Moreover, the stylization task requires keeping the content image structure~\cite{artflow,deng2021stytr2,style_tip}, but the current work of diffusion does not preserve the global content of the input image.
There are several works~\cite{dalle2,latentdiffusion,Huang2022MGAD,disco} that use the content image as the initial input to the diffusion model. 
However, the gradual addition of noise, which can be destructive to the image content, cannot achieve similar stylization applications.
Specifically, DreamBooth~\cite{ruiz2022dreambooth} can maintain the content of a particular subject and embed it into the output domain. However, the method requires fine-tuning for each new set of content images.
In addition, the forward diffusion entails introducing Gaussian noise to the image during the training phase. This noise gradually transforms the image, making it progressively more featureless and resembling a process of entropy increase. As the noise level increases, the order and regularity within the image weaken, leading to significant changes in the content representation.
Therefore it remains challenging to maintain the global content for arbitrary images based on the diffusion model.

\begin{figure*}
    \centering
    \includegraphics[width=\linewidth]{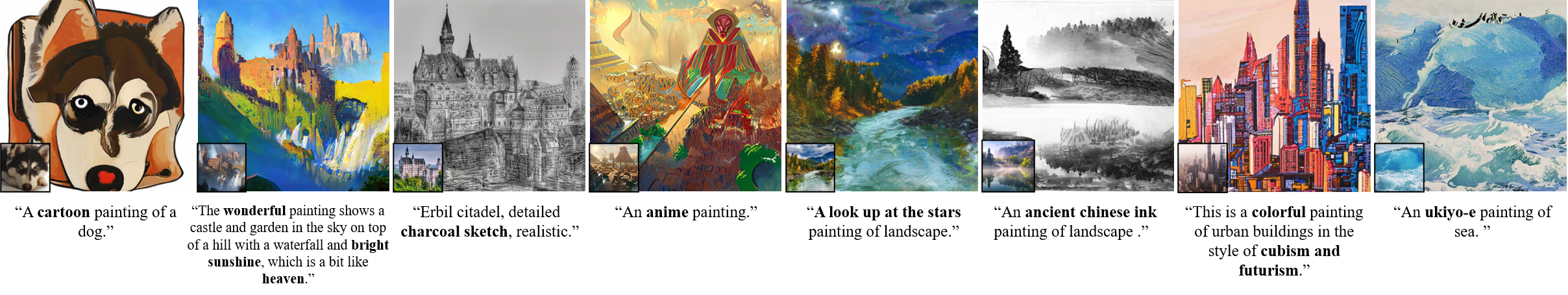}
    \caption{Image stylization results of DiffStyler with diverse text prompts. The input content image for each result is shown as an inset. Using text prompts for stylization, the proposed could generate digital art in various styles.}
    \label{fig:teaser}
\end{figure*}

\IEEEpubidadjcol
To solve the image content preservation problem of the diffusion model, we propose a novel controllable double-diffusion text-driven image stylization method, the working schematic is shown in Fig.~\ref{fig:schema}.
Unlike traditional diffusion models for image generation, we make the following three improvements. First, we replace random noise with learnable noise in the free diffusion process of the content image, preserving the main structure of the content image. Second, for each step of the inference process, we adopt dual diffusion architectures so that the stylized results are both explicit in content and abstract in aesthetics. Furthermore, we optimize the simulation of the diffusion process in terms of numerical methods, so that the generation process can guarantee quality while speeding up the sampling speed. 
Experiments show that our diffusion model-based image content retention and style transfer is effective and achieves excellent results on both quantitative metrics and manual evaluation (see Fig.~\ref{fig:teaser}).
In summary, our main contributions are as follows:
\begin{itemize}
\item We present a new dual diffusion-based text-driven image stylization framework that generates output matching the text prompt of the desired style while preserving the main structure of the input content image.
\item We apply learnable noise based on the content image to overcome the destructive effect of adding random noise to the image content during traditional diffusion.
\item Numerous experimental and qualitative examples show that \shortname outperforms baseline methods and achieves outstanding results with desirable content structures and style patterns.
\end{itemize}


 \section{Related Work}
\label{sec:related_work}
\textbf{Image style transfer.}
Style transfer aims to migrate the style of a painting to a photograph, maintaining its original content.
Initially, Gatys \etal~\cite{gatyscnnstyle} find that hierarchical layers in CNNs can be used to extract image content structures and style texture information and propose an optimization-based iterative method to generate stylized images. 
An increasing number of methods have been developed thereafter to advance the quality of stylization. 
Arbitrary style transfer in real-time~\cite{johnson2016perceptual} is improved by minimizing perceptual loss which is the combination of feature reconstruction loss as well as the style reconstruction loss. 
More generally, arbitrary style transfer has gained more attention in recent years.

Huang \etal~\cite{huangadain} propose an adaptive instance normalization (AdaIN) to replace the mean and variance of content with that of style. AdaIN is widely adopted in image generation tasks~\cite{StyleGAN,wang2020collaborative} to fuse the content and style features. 
Deng \etal~\cite{deng2021stytr2} propose StyTr$^2$ which contains two different transformer encoders to generate domain-specific sequences for content and style, respectively. 
Zhang \etal~\cite{Zhang:2022:CAST} present contrastive arbitrary style transfer (CAST) to learn style representation directly from image features by analyzing the similarities and differences between multiple styles and taking the style distribution into account.
In conclusion, while traditional image style transfer is stable, it requires the provision of additional style images and the results lack artistry and creativity.

\textbf{Text-driven image manipulation.}
In the existing text-guided image synthesis~\cite{huang2023region,zhang2023prospect}, the encoders for text embedding work as guide conditions for generative models.
OpenAI proposes the high-performance text-image embedding model CLIP~\cite{clip}, based on which several methods manipulate images with textual conditions.
StyleCLIP~\cite{StyleCLIP} performed attribute manipulation by exploring the learned latent space of StyleGAN~\cite{StyleGAN}. 
They control the generative process toward a given textual condition by finding the appropriate vector direction.
Therefore, StyleGAN-NADA~\cite{StyleGANNADA} proposed a model modification method using text conditions only and modulated the trained model into a novel domain.
Although these methods have been successful in specific domains, they are difficult to apply to arbitrary data.

Based on the above works, Kwon \etal~\cite{CLIPstyler} proposed CLIPstyler that contains a patch-wise text-image matching loss with multiview augmentations for realistic texture transfer.
Fu \etal~\cite{fu2022ldast} propose a contrastive language visual artist (CLVA) that learns to extract visual semantics from style instructions and accomplish LDAST by the patch-wise style discriminator.
The stylization effect of CLIPstyler~\cite{CLIPstyler} and LDAST~\cite{fu2022ldast} is dependent on the patch style discriminator, and the quality of the randomly sampled patch will have a critical impact on the results of the transfer.

\textbf{Diffusion models for image synthesis.}
The diffusion model~\cite{2015Diffusion} is a generative model learning to generate images by removing noise from random signals in a stepwise manner, which has received a great deal of attention recently.
Sohl-Dickstein \etal~\cite{2015Diffusion} were the first to implement image generation using the diffusion model, and with continued research in subsequent approaches~\cite{YangSong2019GenerativeMB,Guided-Diffusion,DDPM,DiffusionCLIP,nichol2021glide,dalle2,imagen}, the diffusion model is generating high-resolution images with unprecedented quality, often surpassing GANs.

Although denoising diffusion probability models (DDPMs)~\cite{DDPM} can produce high-quality samples, they require hundreds to thousands of iterations to produce the final samples. Some previous methods have successfully accelerated DDPMs by adjusting variance schedules (e.g., IDDPMs~\cite{IDDPM}) or denoising equations (e.g., DDIMs~\cite{DDIM}). However, these acceleration methods cannot maintain the quality of the samples. Liu \etal~\cite{pndm}  proposed the idea that DDPMs should be considered as solving differential equations on manifolds and proposed pseudo-numerical methods for diffusion models (PNDMs) to accelerate the inference process while maintaining the sample quality.

Several large-scale text-image models have recently emerged, such as DALL·E 2~\cite{dalle2}, GLIDE~\cite{nichol2021glide}, and Imagen~\cite{imagen}, demonstrating unprecedented image generation results.
Notably, some studies have focused on enhancing the control of the progressive inference process~\cite{Guided-Diffusion}, thereby endowing diffusion models with remarkable controllability. Noteworthy advancements in the field of style transfer have been realized through the remarkable text-to-image diffusion models developed by GLIDE~\cite{nichol2021glide} and stable diffusion~\cite{latentdiffusion}, enabling diffusion-based approaches to achieve superior outcomes.

Since the above works are not designed to maintain the structure of the input image, this does not quite satisfy the problem setting of traditional style transfer tasks~\cite{gatyscnnstyle,tnnls_style3}.
If the above works~\cite{latentdiffusion, dalle2} are applied to an image-to-image application with additional steps, the results will differ significantly from the input image. Kim \etal~\cite{DiffusionCLIP} introduced the method known as DiffusionCLIP, which focuses on performing global changes in images. However, it should be noted that DiffusionCLIP is limited in its applicability to specific domains, and is not designed to operate on arbitrary images. In contrast to existing approaches, our objective is to develop a novel method that enables the stylization of arbitrary images guided by text prompts through the utilization of the diffusion method. By leveraging the power of diffusion, we aim to provide users with precise and controllable stylization capabilities, expanding beyond the constraints of specific domains.

\begin{figure*}
    \centering
    \includegraphics[width=\linewidth]{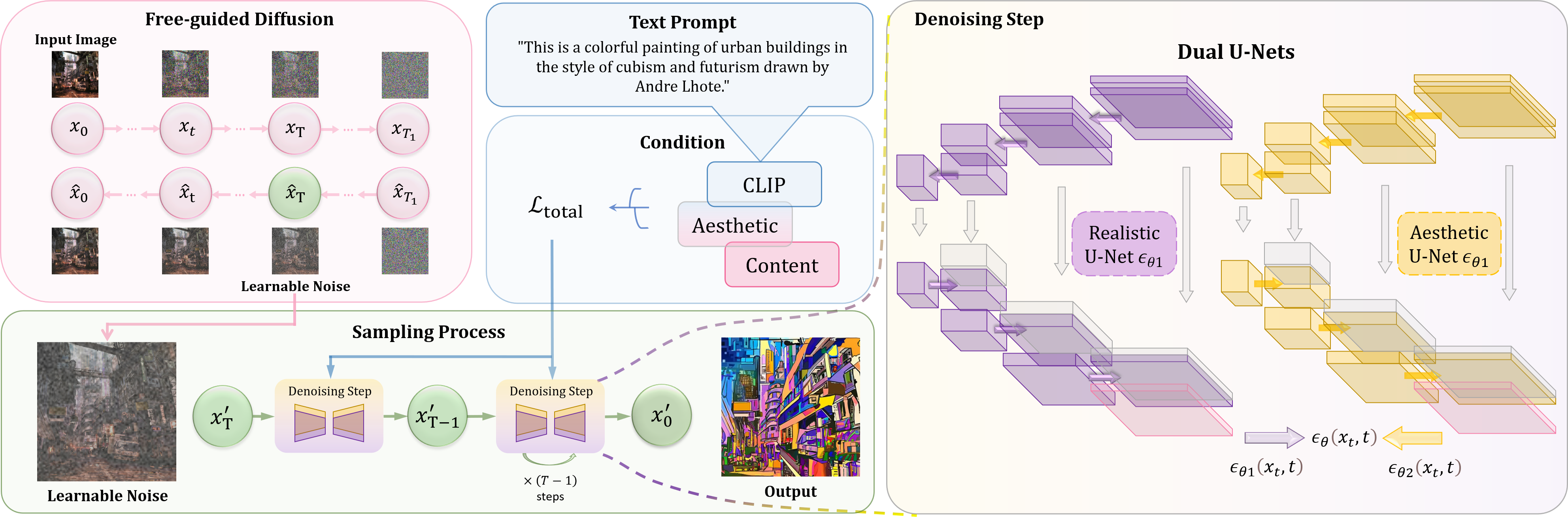}
    \caption{The pink module represents the free-guided diffusion process, which generates the learnable noise denoted as $\hat{\textbf{x}}_{\mathrm{T}}$. 
    The $\hat{\textbf{x}}_{\mathrm{T}}$ is set to be equal to $\textbf{x}_{\mathrm{T}}^{\prime}$ as the starting map for the sampling process of the green module. 
    In addition, the guided condition of the blue module is incorporated into the denoising step of the sampling process. The dual denoising U-Nets architecture, represented by the yellow and purple mixed-color module, is employed for each step. The purple architecture corresponds to the realistic U-Net, while the yellow architecture represents the aesthetic U-Net. After $\mathrm{T}$ steps of denoising, the final output $\textbf{x}_{0}^{\prime}$ is obtained.}
    \label{fig:pipeline}
\end{figure*}

\textbf{Noise impact.}
Given the uniqueness of the effect of noise on the diffusion process, many researchers~\cite{videocontrolnet,pfbdiff,videofusion} have conducted in-depth works on input noises.
For instance, Video ControlNet~\cite{videocontrolnet} proposes that the initial noisy samples used in the denoising process have a significant impact on texture synthesis.
Since spatial translation or distortion of the noise can change the output semantics, Video ControlNet introduces a well-designed input noise combination scheme. By employing a sliding window approach to maximize the temporal consistency of the generated video. 
However, discrepancies and inaccuracies in the estimation of the optical flow result in some temporal discontinuities remaining in the generated video.

PFB-Diff~\cite{pfbdiff} utilizes initial random noise and combines progressive feature blending and attention masking mechanisms to generate intermediate noisy images, which are then progressively denoised to obtain the final edited image.
One limitation is that the model may struggle to generate desired scenes in background replacement. Another limitation is that the ability to describe desired objects through text remains limited, making personalized editing challenging. 

More recently, VideoFusion~\cite{videofusion} resolves the per-frame noise into two parts, namely base noise and residual noise, where the base noise is shared by consecutive frames. 
While VideoFusion sharing the underlying noise between consecutive frames helps to utilize temporal correlation better, it also limits the motion in the generated video and does not apply to video generation with large differences between frames.
This allows the image priors of the pre-trained model to be efficiently shared by all frames and thereby facilitate the learning of video data.

In contrast to the aforementioned research, our study investigates the influence of noise on content preservation within the domain of style transfer. In addition, an attempt is made to employ learnable noise within a diffusion modeling framework to facilitate a style stylization task.

\section{Method}
\label{sec:method}
We now formally introduce \shortname, a controllable dual diffusion framework for text-driven image stylization. Given a content image $\textbf{x}_0$ and the target text prompt $\textprompt$, \shortname can transfer $\textbf{x}_0$ into a stylized one with the desired style. As illustrated in Fig.~\ref{fig:pipeline}, \shortname consists of three main stages.
First, we input the content image $\textbf{x}_{0}$ into the diffusion model and let it perform the $T_1$ steps of free-guided diffusion to obtain learnable noise. 
Second, the result $\hat{\textbf{x}}_{T}$ at the $T$ steps of free-guided diffusion is used as the input $\textbf{x}_{T}^{\prime}$ to the dual diffusion model for the reverse sampling process. 
Third, in the reverse sampling process, we perform the relevant optimization.
The process is depicted in Algorithms~\ref{alg:alg1} and \ref{alg:alg2}.
We investigate the inherent properties of the diffusion model and propose novel content-preserving learnable noise and a dual diffusion pathway for style/content decoupling, and successfully enable the diffusion model to break through the limitations of the diffusion model in style transfer tasks.


\subsection{Dual Diffusion Models}
\textbf{Denoising Diffusion Probabilistic Models.}
Recently, denoising diffusion probability models (DDPMs) have been shown to generate high-quality images~\cite{Guided-Diffusion,DDPM,dalle2}. DDPMs learn the denoising process for parametrized Markov noise images. The isotropic Gaussian noise samples are converted to samples from the training distribution. In the following, we provide a brief overview of DDPMs~\cite{Guided-Diffusion,DDPM,DDIM,IDDPM}.

At the core of DDPMs lies the forward noising process, which involves adding Gaussian noise with variance $\beta_t \in(0,1)$ at each time step $t$ to an initial data distribution $\textbf{x}_0 \sim q\left(\textbf{x}_0\right)$. This creates a sequence of images $\textbf{x}_1, \ldots, \textbf{x}_T$ governed by the following equations:
\begin{equation}
\begin{aligned}
q\left(\textbf{x}_1, \ldots, \textbf{x}_T \mid \textbf{x}_0\right) & =\prod_{t=1}^T q\left(\textbf{x}_t \mid \textbf{x}_{t-1}\right), \\
q\left(\textbf{x}_t \mid \textbf{x}_{t-1}\right) & =\mathcal{N}\left(\sqrt{1-\beta_t} \textbf{x}_{t-1}, \beta_t \mathbf{I}\right),
\end{aligned}
\end{equation}
where $q\left(\textbf{x}_1, \ldots, \textbf{x}_T \mid \textbf{x}_0\right)$ denotes the joint distribution of the image sequence given the initial image $\textbf{x}_0$, and $q\left(\textbf{x}_t \mid \textbf{x}_{t-1}\right)$ represents the conditional distribution of image $\textbf{x}_t$ given the previous image $\textbf{x}_{t-1}$. It is important to note that as the number of steps $T$ increases, the final output $\textbf{x}_T$ tends to approximate an isotropic Gaussian distribution.

A key feature of the forward noising is that each step $\textbf{x}_t$ may be sampled straight from $\textbf{x}_0$, eliminating the need to construct intermediary stages.
\begin{equation}
\begin{aligned}
\label{E2}
q\left(\textbf{x}_t \mid \textbf{x}_0\right) & =\mathcal{N}\left(\sqrt{\bar{\alpha}_t} \textbf{x}_0,\left(1-\bar{\alpha}_t\right) \mathbf{I}\right), \\
\textbf{x}_t & =\sqrt{\bar{\alpha}_t} \textbf{x}_0+\sqrt{1-\bar{\alpha}_t} \epsilon,
\end{aligned}
\end{equation}
where $\epsilon \sim \mathcal{N}(\textbf{0}, \mathbf{I})$ denotes a Gaussian noise sample, $\alpha_t=1-\beta_t$ represents the complementary noise level at time $t$, and $\bar{\alpha}_t=\prod_{s=0}^t \alpha_s$ captures the accumulated noise level up to time $t$. Importantly, this process can be inverted to generate a fresh sample from the distribution $q\left(\textbf{x}_0\right)$.
The Markovian process is reversed to generate a fresh sample from the distribution $q\left(\textbf{x}_0\right)$. 
The posteriors, $q\left(\textbf{x}_{t-1} \mid \textbf{x}_t\right)$, which were demonstrated to also be Gaussian distributions~\cite{2015Diffusion}.
They are sampled beginning from a Gaussian noise sample, $\textbf{x}_T \sim \mathcal{N}(\textbf{0}, \mathbf{I})$, to produce a reverse sequence.
$q(\textbf{x}_{t-1} \mid \textbf{x}_t)$ is determined by the unknown data distribution $q(\textbf{x}_0)$.

To predict the mean and covariance of $\textbf{x}_{t-1}$ using the input $\textbf{x}_t$, a deep neural network $p_\theta$ is employed. This network enables the sampling of $\textbf{x}_{t-1}$ from a normal distribution parameterized by these statistics:
\begin{equation}
p_\theta\left(\textbf{x}_{t-1} \mid \textbf{x}_t\right)=\mathcal{N}\left(\mu_\theta\left(\textbf{x}_t, t\right), \Sigma_\theta\left(\textbf{x}_t, t\right)\right).
\end{equation}
Since it is very difficult to infer $\mu_\theta\left(\textbf{x}_t, t\right)$ directly, we refer to~\cite{DDPM} and first compute the noise prediction $\epsilon_\theta\left(\textbf{x}_t, t\right)$ and then substitute it into Bayes' theorem (refer to Equation~\ref{E2}) to derive: 
\begin{equation}
\mu_\theta\left(\textbf{x}_t, t\right)=\frac{1}{\sqrt{\alpha_t}}\left(\textbf{x}_t-\frac{\beta_t}{\sqrt{1-\bar{\alpha}_t}} \epsilon_\theta\left(\textbf{x}_t, t\right)\right).
\end{equation}

\textbf{Pseudo numerical methods for DDPM.}
Diffusion models convert Gaussian data to images iteratively. They require many iterations for high-quality samples, slowing large sample generation. Numerical methods have limitations within a limited range. 
Another problem arises when using numerical methods with the diffusion model equation. The neural network and the equation are well-defined only within a limited range. To address this, a reverse process is used to calculate the derivative of the generated data.

The diffusion model equation is unbounded in most cases, which means that the generation process can produce samples far away from the well-defined area, introducing new errors.
The related differential equations of the diffusion model may be derived directly and self-consistently to provide a theoretical relationship between diffusion processes and numerical methods:
\begin{equation}
\label{E1}
\frac{d \textbf{x}}{d t}=-\bar{\alpha}^{\prime}(t)\left(\frac{\textbf{x}(t)}{2 \bar{\alpha}(t)}-\frac{\epsilon_\theta(\textbf{x}(t), t)}{2 \bar{\alpha}(t) \sqrt{1-\bar{\alpha}(t)}}\right).
\end{equation}

The equation of the numerical method is redefined as: 
\begin{equation}
\textbf{x}_{t+\delta}=\textbf{x}_t+\delta f \Rightarrow \textbf{x}_{t+\delta}=\phi\left(\textbf{x}_t, \epsilon_t, t, t+\delta\right).
\end{equation}
Here, $f$ is Equation~\ref{E1}.
It has been experimentally demonstrated~\cite{pndm} that the pseudo-linear multi-step method is the most efficient method for the diffusion model with similar generation quality.

\begin{figure*}[thbp]
    \centering
    \includegraphics[width=\linewidth]{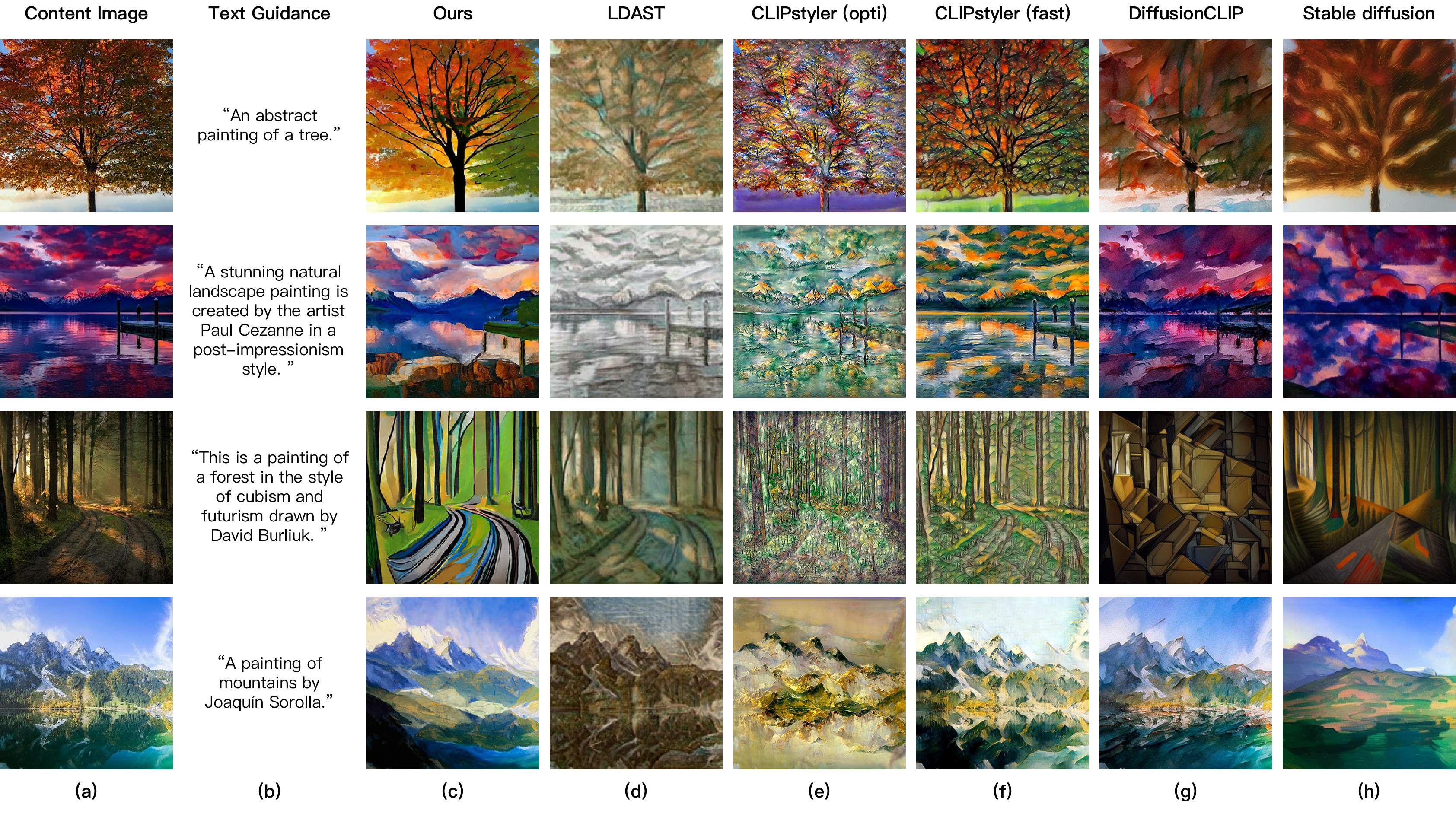}
    \caption{Qualitative comparison with state-of-the-art image style transfer methods including LDAST, CLIPstyler (opti), CLIPstyler (fast), DiffusionCLIP, and Stable Diffusion.}
    \label{fig:comparison}
\end{figure*}

\begin{algorithm}[!t]
\caption{Learnable noise generating, given \\ a diffusion model $(\mu_{\theta}(\textbf{x}_{t}), \Sigma_{\theta}(\textbf{x}_{t}))$.}\label{alg:alg1}
\begin{algorithmic}
\STATE 
\REQUIRE The input image $\textbf{x}_0$, free-guided noising steps $T_1$, denoising steps $T$.
\ENSURE Learnable noise $\hat{\textbf{x}}_{T}$.
\STATE $t = {T}_{1}$
\STATE $\textbf{x}_{T_1} \leftarrow$ sample from $\mathcal{N}(\textbf{0}, \mathbf{I})$
\STATE $\hat{\textbf{x}}_{T_1} = {\textbf{x}}_{T_1}$
\REPEAT
    \STATE $t - 1 \gets t$ 
    \STATE $\mu_{\theta}(\hat{\textbf{x}}_{t}) \leftarrow \boldsymbol{\epsilon}_\theta(\hat{\textbf{x}}_t, t \mid \emptyset)$
    \STATE $\mu, \Sigma \leftarrow \mu_{\theta}(\hat{\textbf{x}}_{t}), \Sigma_{\theta}(\hat{\textbf{x}}_{t})$
    \STATE $\hat{\textbf{x}}_{t-1} \sim \mathcal{N}\left(\mu, \Sigma\right)$
\UNTIL {$t = T$}   
\end{algorithmic}
\label{alg1}
\end{algorithm}
\textbf{Learnable noise.}
The diffusion model~\cite{2015Diffusion} is inspired by nonequilibrium thermodynamics.
They define a Markov chain of diffusion steps that adds random noise slowly to the data and then learns to reverse the diffusion process by reconstructing the desired data samples from the noise.
In previous diffusion work, either a random Gaussian noise image or an input image superimposed with a corresponding amount of random Gaussian noise is usually used as the starting image $\textbf{x}_{T}^{\prime}$ for the sampling process. This operation does not protect the content structure of the input image, making the generated image differ significantly from the input image in terms of content, as shown in the results of stable diffusion in Fig.~\ref{fig:comparison}.
Despite the remarkable progress of diffusion models for image generation, the destructive nature of the stochastic noising process and the random nature of the reverse denoising process make it difficult to preserve the content of the original image, leading that style transfer by diffusion methods still not well explored. 
In this context, we investigate and exploit the inherent properties of the diffusion model. Our findings indicate that the final denoising outcome of $\textbf{x}_T$ is achieved by performing $T$ steps of free diffusion on the input image without any textual guidance. Specifically, we introduce a zero embedding as a guiding condition for the denoising network, represented by
\begin{equation}
\boldsymbol{\epsilon}_\theta(\textbf{x}_t, t)=\boldsymbol{\epsilon}_\theta(\textbf{x}_t, t \mid \emptyset). 
\end{equation}
The denoising network $\epsilon_\theta(\textbf{x}_0)$ represents the parametrized conditional model.
Typically, the number of forward and sampling process steps in diffusion methods are equal. However, we discover that increasing the number of free diffusion steps $T_1$ beyond the number of reverse diffusion steps $T$ better preserves the content structure of the input image (as observed in Fig.~\ref{fig:learnablenoise}). 
Therefore, we adopt $\hat{\textbf{x}}_{T}$ as the initial image $\textbf{x}_{T}^\prime$ for the sampling process in our experiments, as shown by the pink arrow in pipeline Fig.~\ref{fig:pipeline}. $T_1$ and $T$ are set to $150$ and $50$, respectively. This approach ensures enhanced preservation of the content structure of the input image, rendering it suitable for image stylization tasks.
This process is summarized in Algorithms \ref{alg:alg1} and \ref{alg:alg2}.

\textbf{Basic architecture of DiffStyler.}
We found that if the diffusion model is trained only on the natural image dataset, the results will be close to real images and lack artistic appearance. 
If the diffusion model is trained only on the artistic image training set, the results will be too abstract to preserve the input content.
Therefore, we used light denoising networks $\boldsymbol{\epsilon}_{\theta 1}$ and $\boldsymbol{\epsilon}_{\theta 2}$ trained on the Conceptual 12M~\cite{changpinyo2021cc12m,modelckpt} and WikiArt~\cite{WikiArt} datasets, respectively.
Then, we perform sampling using the following linear combination of the dual channel score estimates:
\begin{equation}
\label{dual}
\tilde{\boldsymbol{\epsilon}}_\theta\left(\textbf{x}_t, t \mid \textprompt \right)= w \boldsymbol{\epsilon}_{\theta 1} \left(\textbf{x}_t, t \mid \textprompt \right)+(1-w) \boldsymbol{\epsilon}_{\theta 2} \left(\textbf{x}_t, t \mid \textprompt \right),
\end{equation}
where $0<w<1$.
We utilize it for each step of the inference process, resulting in stylized outputs that are both explicit in content and abstract in aesthetics.

\begin{algorithm}[!t]
\caption{Text guided stylized diffusion sampling, given \\ a diffusion model $(\mu_{\theta}(\textbf{x}_{t}), \Sigma_{\theta}(\textbf{x}_{t}))$.}\label{alg:alg2}
\begin{algorithmic}
\STATE 
\REQUIRE The input image $\textbf{x}_0$, text prompt $\textprompt$, gradient scale $w$, free-guided noising steps $T_1$, denoising steps $T$.
\ENSURE Stylized result $\textbf{x}_{0}^{\prime}$ guided by $\textprompt$.
\STATE $t = T$
\STATE $\hat{\textbf{x}}_{T} = $ learnable noise ($\textbf{x}_{T}^{\prime}$)
\STATE $\textbf{x}_{T}^{\prime} = \hat{\textbf{x}}_{T}$
\STATE \textbf{for} t = T, ... , 1 \textbf{do}
    \STATE \quad$\mu, \Sigma \leftarrow \mu_{\theta}(\textbf{x}_{t}^{\prime}), \Sigma_{\theta}(\textbf{x}_{t}^{\prime})$
    \STATE \quad$\tilde{\boldsymbol{\epsilon}}_{\theta}( \textbf{x}_{t}^{\prime}, t \mid \textprompt) \leftarrow  w \cdot {\epsilon _{\theta_1 }} ( \textbf{x}_{t-1}^{\prime}, t \mid \textprompt )+ (1-w) \cdot\epsilon_{\theta_2}(\textbf{x}_{t}^{\prime}, t \mid \textprompt) $
    \STATE \quad$\mu_{\theta}(\textbf{x}_{t}^{\prime}) \leftarrow \tilde{\boldsymbol{\epsilon}}_{\theta}( \textbf{x}_{t}^{\prime}, t \mid \textprompt)$
    \STATE \quad$\mathcal{L}_{total} \gets \lambda_d \mathcal{L}_{inst }+\lambda_{c1} \mathcal{L}_{c}+\lambda_{c2} \mathcal{L}_{c_{patch}}+\lambda_{aes} \mathcal{L}_{aes}+$
    \STATE \qquad \qquad \quad$\lambda_{t v} \mathcal{L}_{t v}$
    \STATE \quad$\textbf{x}_{t-1}^{\prime} \leftarrow$ sample from $\mathcal{N}(\mu_{\theta}(\textbf{x}_{t}^{\prime})+\Sigma\mathcal{L}_{total}, \Sigma)$

\STATE \textbf{end for}
\RETURN $\textbf{x}_{0}^{\prime}$
\end{algorithmic}
\label{alg2}
\end{algorithm}

\subsection{Network Optimization}
\textbf{Instruction loss.}
We leverage a pre-trained ViT-B/16 CLIP~\cite{clip} model to stylize the content image according to the text prompt.
The cosine distance between the CLIP embedding of the transferred image $\textbf{x}_t$ during the diffusion process and the CLIP embedding of the text prompt $\textprompt$ may be used to specify the CLIP-based loss, or $\mathcal{L}_{inst}$.
We define the language guidance function using the cosine distance, which measures the similarity between the embeddings $E_{\textbf{x}_t}$ and $E_{\textprompt}$.
The text guidance function can be defined as:
\begin{equation}
\mathcal{L}_{inst}\left(\textbf{x}_{t}, t, \textprompt\right)=\mathcal{D}_{CLIP}(E_{\textbf{x}_t} , E_{\textprompt}),
\end{equation}
where $\mathcal{D}_{CLIP}$ is the cosine distance of the CLIP embeddings.

\textbf{Content perceptual loss.}
To further enhance the alignment with the content of input images, we use feature maps extracted to compute the content loss and patchwise contrastive content loss~\cite{gatyscnnstyle,park2020contrastive}:
\begin{equation}
\mathcal{L}_c=\frac{1}{N_l} \sum_{i=0}^{N_l}\left\|\phi_i\left(\textbf{x}_{0}^{\prime}\right)-\phi_i\left(\textbf{x}_{0}\right)\right\|_2,
\end{equation}
where $\phi_i(\cdot)$ denotes features extracted from the $i$-th layer in a pre-trained VGG19 and ${N_l}$ is the number of layers.

Besides, we utilize contrast learning~\cite{park2020contrastive} to match corresponding input-output patches at a specific location. We can leverage the other patches within the input as negatives.
Let $\textbf{x}_{0}^{\prime}, \textbf{x}_{0} \in \mathbb{R}^{H_{patch} \times W_{patch} \times C}$ be patched from the two signals, in the same spatial location. 
The patch with resolution $H \times W$ is an RGB image patch ($C = 3$).
The encoder $F$ is a simple linear projection of a patch of pixels.
The embedding vectors of the network generated and input image patches are defined as $v=F\left(\textbf{x}_{0}^{\prime}\right)$ and $v^{+}=F\left(\textbf{x}_{0}\right)$, respectively.
The $N$ negatives are mapped to $K$-dimensional vectors and $\boldsymbol{v}_n^{-}$ denotes the $n$-th negative.
The loss is specified as an $(N+1)$-way classification problem with logits proportional to the similarity of embedded patches.
To represent the probability of a positive example being selected over a negative example, the cross-entropy loss is calculated as follows:
\begin{equation}
\begin{split}
&\ell_{c_{patch}}(v, v^{+}, v^{-})=\\
&-\log \left[\frac{\exp (s\left({v} \cdot {v}^{+}) / \tau\right)}{\exp (s\left({v} \cdot {v}^{+}) / \tau\right)+\sum_{n=1}^N \exp (s\left({v} \cdot {v}_n^{-}) / \tau\right)}\right],
\end{split}
\end{equation}
where $\tau=0.07$ is a parameter to control temperature. $s\left(v_1, v_2\right)=v_1^{\mathrm{T}} v_2$ delivers the dot product of the similarity of two encoded patch signals.
We leverage the notation $v_l^s \in \mathbb{R}^{D_l}$ to index into the tensor, which is the ${D_l}$-dimensional feature vector at the $s^{th}$ spatial location.
The collection of feature vectors at all other spatial locations is represented as $\bar{v}_l^s \in \mathbb{R}^{\left(S_l-1\right) \times D_l}$, where $S_l$ is the number of spatial locations of the tensor.
And the patchwise contrastive content loss is defined as
\begin{equation}
\mathcal{L}_{c_{patch}}(\textbf{x}_{0}^{\prime}, \textbf{x}_{0})=\sum_{l=1}^L \sum_{s=1}^{S_l} \ell_{c_{patch}}\left(\hat{v}_l^s, v_l^s, \bar{v}_l^s\right).
\end{equation}

\textbf{Aesthetic loss.}
We also employ an aesthetic loss to make the model's representation of style more consistent with human preferences.
We leverage a model to fit and inference code for CLIP aesthetic regressions $R$ trained on Simulacra Aesthetic Captions~\cite{aescrowson,aesdavid}.
It is a dataset of over 238,000 synthetic images generated by diffusion models as well as human ratings.
We use it as a scorer to evaluate the results generated by \shortname, and return the weighted aesthetic exploring loss for optimization:
\begin{equation}
\mathcal{L}_{aes}= - R\left(\textbf{x}_t, \textprompt \right).
\end{equation}
$\mathcal{L}_{aes}$ improves the overall visual quality.

\textbf{Total variation loss.}
Total variation loss ($\mathcal{L}_{t v}$) serves as a regularization technique extensively utilized in image processing and computer vision applications. Its primary objective is to promote the presence of smooth transitions and minimize noise by penalizing sudden alterations in intensity or gradients.
The mathematical representation for the $\mathcal{L}_{t v}$ is defined as:
\begin{equation}
\mathcal{L}_{t v} = \sum_{i,j} \left( \left| I_{i+1,j} - I_{i,j} \right| + \left| I_{i,j+1} - I_{i,j} \right| \right)
\end{equation}

In the equation, the symbol $I$ corresponds to the image under consideration, while $I_{i,j}$ denotes the pixel intensity located at the spatial coordinates $(i,j)$. This loss computation involves the absolute differences between neighboring pixel intensities in both the horizontal and vertical directions. By summing up these differences across the entire image, the $\mathcal{L}_{t v}$ encourages smoothness by discouraging abrupt variations in pixel values.

As a consequence, the total loss is formulated as follows:
\begin{equation}
\label{Eq_loss}
\mathcal{L}_{total}=\lambda_d \mathcal{L}_{inst }+\lambda_{c1} \mathcal{L}_{c}+\lambda_{c2} \mathcal{L}_{c_{patch}}+\lambda_{aes} \mathcal{L}_{aes}+\lambda_{tv} \mathcal{L}_{tv}.
\end{equation}
We set $\lambda_d$, $\lambda_{c1}$, $\lambda_{c2}$, $\lambda_{aes}$, and $\lambda_{tv}$ to $50$, $3.0$, $1.0$, $10$, and $80$, respectively. 
The impacts of loss weights are discussed in Section~\ref{subsec:ablation}.

\begin{figure*}
    \centering
    \includegraphics[width=\linewidth]{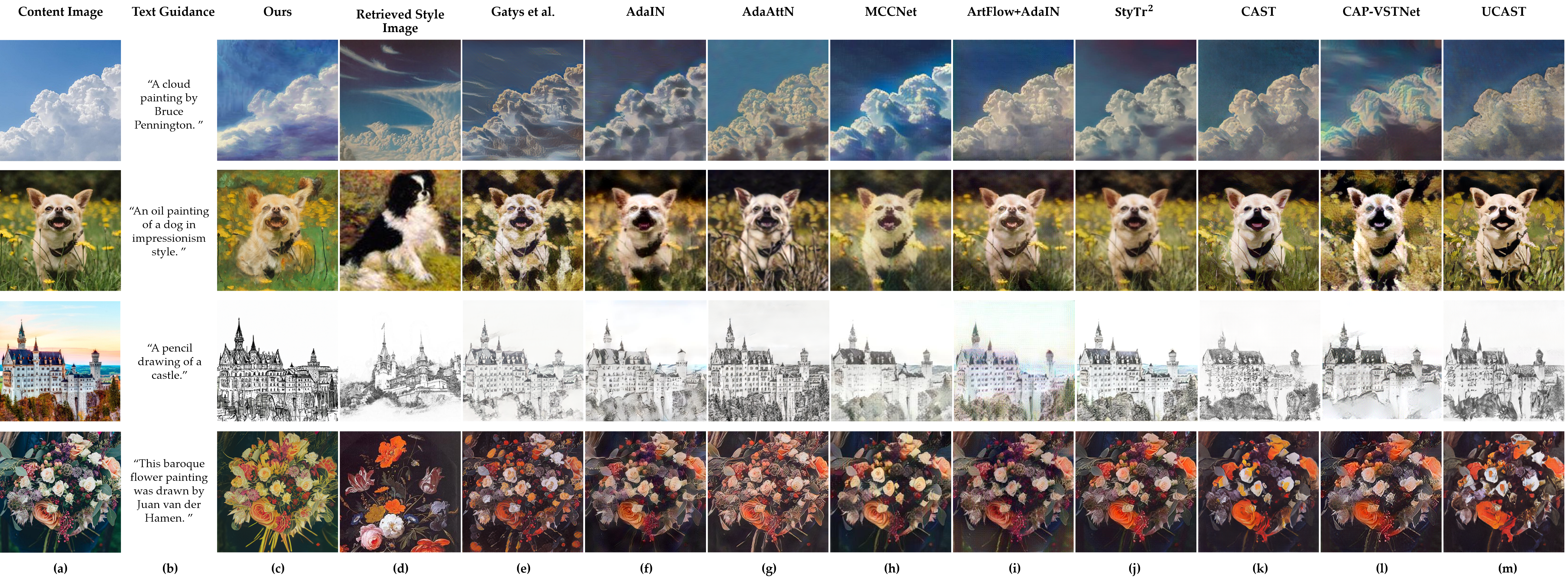}
    \caption{Qualitative comparison with image style transfer methods. Our approach uses the text guidance directly as the style condition, while others need to first retrieve a style reference from an image database based on the text input before generating the style transfer results.}
    \label{fig:comparison2}
\end{figure*}

\section{Experiments}

\subsection{Implementation Details}
For the CLIP~\cite{clip} model, we use ViT-B/16 as the vision transformer~\cite{vision-transformer}.
For the diffusion model, we use the primary model~\cite{modelckpt} of a resolution $256 \times 256$ trained on the Conceptual 12M~\cite{changpinyo2021cc12m} dataset.
In addition, we use a pre-trained model~\cite{modelckpt} on the WikiArt~\cite{WikiArt} dataset as a secondary diffusion model to enable a superior artistic representation of the entire model.
Our approach to the combination of the two pre-trained models can be extended to the combination of any diffusion model according to the demands of the application scenario. This results in significant savings in training resources.
The output resolution of \shortname is $256 \times 256$. 
The U-Net~\cite{UNet} architecture we utilize is based on Wide-ResNet~\cite{wideres}. 
To ensure the quality of the results and to maintain consistency of the parameters, the diffusion step and the time step used for the experiments in our work are set to $50$.

\subsection{Computing Consumptions}
DiffStyler takes $18$ seconds to generate a $256\times256$ image on a single NVIDIA GeForce RTX $3090$. 
It takes $14$ seconds with a single channel. 
This is a moderate amount of time compared to related work.
DiffusionCLIP~\cite{DiffusionCLIP} and Stable diffusion~\cite{latentdiffusion} are two other diffusion-based methods that take significantly longer to reason about compared to our model. This difference is because they require longer inference steps to obtain relatively desirable results. Although there is still a gap between our current speed compared to the non-denoising method~\cite{fu2022ldast}, given the continuous progress in accelerating the denoising inference process, it can be expected that the efficiency of the current inference process will be further improved in the foreseeable future.

The inference process for one image consumes about 4 GB (6 GB and 16 GB for DiffusionCLIP~\cite{DiffusionCLIP} and Stable Diffusion~\cite{latentdiffusion}, respectively), which is much smaller than the general diffusion model~\cite{ruiz2022dreambooth,imagen}. 
The WikiArt model and the CC12M model, trained using 4 NVIDIA GeForce RTX 3090s, required approximately 40 minutes and 50 hours of training time, respectively.


\subsection{Qualitative Evaluation}
\paragraph{Comparison with text-driven stylization}
We compare DiffStyler with SOTA text-driven image stylization methods including LDAST~\cite{fu2022ldast}, CLIPstyler~\cite{CLIPstyler}, DiffusionCLIP~\cite{DiffusionCLIP}, and Stable Diffusion~\cite{latentdiffusion}. There are two versions of CLIPstyler. CLIPstyler (opti) requires real-time optimization on each content and each text. CLIPstyler (fast) requires real-time optimization on each text.
We can observe that LDAST has relatively similar style performance results in different text-driven cases and is less artistic.
CLIPstyler (opti) texture patches used to convey style are substantially corrupting the content of the input image, e.g. (the 3$^{rd}$ row) the dog's eyes are shifted to other locations in the result.
Both CLIPstyler (opti) and CLIPstyler (fast) are more focused on details and lack macro lines, color blocks, and textures.
DiffusionCLIP and Stable Diffusion do not express style as prominently (e.g., the 1$^{st}$, 2$^{nd}$, and 4$^{th}$ rows) and result in a severe content loss (e.g., the 3$^{rd}$ row) under some textual conditions.


\begin{table*}
  \caption{Quantitative comparisons for text-driven image style transfer. We compute the average learned perceptual image patch similarity (LPIPS) and clip score. The best results are highlighted in \textbf{bold} while the second best results are marked with an \underline{underline}.}
  \label{tab:quantity}
  \centering
  \begin{tabular}{l|ccccccc}
    \toprule
         & Ours & LDAST~\cite{fu2022ldast}  & CLIPstyle (opti)~\cite{CLIPstyler} & CLIPstyle (fast)~\cite{CLIPstyler}  & DiffusionCLIP~\cite{DiffusionCLIP}  & Stable diffusion~\cite{latentdiffusion}   \\
    \midrule
    Inference time (s/image) ↓ & \underline{18}  & \textbf{1} & 20 & 45 & 40  &20 \\
    \midrule
    CLIP score~\cite{clip} ↑ & \textbf{0.2869}  & 0.2235 & 0.2422 & 0.2515 & 0.2389  &\underline{0.2773} \\
    \midrule
    LPIPS~\cite{LPIPS} ↓ & \underline{0.505}  & 0.744 & 0.573 & \textbf{0.472} & 0.619 & 0.637 \\
    \bottomrule
  \end{tabular}
\end{table*}

\begin{table*}[htbp]
  \centering
  \caption{Quantitative comparison with state-of-the-art traditional stylization methods. The best results are shown in \textbf{bold}, while the second-best results are marked with an \underline{underline}.}
    \resizebox{\textwidth}{!}{\begin{tabular}{l|ccccccccccccc}
    \toprule
                & Ours        & Gatys et al.~\cite{gatyscnnstyle}      & AdaIN~\cite{huangadain}        & AdaAttN~\cite{liu2021adaattn}    & MCCNet~\cite{mccnet_video}    & ArtFlow+AdaIN~\cite{artflow}       & StyTr$^2$~\cite{deng2021stytr2}       & CAST~\cite{Zhang:2022:CAST}       & CAP-VSTNet~\cite{CAP-VSTNet}       & UCAST~\cite{ucast} \\
    \midrule
    content loss ↓ & \underline{0.156}     & 0.188       & 0.176       & 0.162        & 0.159        & 0.172        & \textbf{0.148}         & \textbf{0.148}       & 0.175       & 0.158 \\
    \bottomrule
    \end{tabular}}
  \label{tab:quantity_tra}%
\end{table*}%

\paragraph{Comparison with style transfer}
We indirectly compare our results with existing style transfer methods, by comparing the proposed DiffStyler with the ``retrieval-stylization'' baselines (see Fig.~\ref{fig:comparison2}). 
The text retrieved art images are firstly used as style images~\cite{clip-retrieval} and then stylized using the state-of-the-art available artistic style transfer methods, including Gatys \etal~\cite{gatyscnnstyle}, AdaIN~\cite{huangadain}, AdaAttN~\cite{liu2021adaattn}, MCCNet~\cite{mccnet_video}, ArtFlow+AdaIN~\cite{artflow}, StyTr$^2$~\cite{deng2021stytr2},  CAST~\cite{Zhang:2022:CAST}, CAP-VSTNet~\cite{CAP-VSTNet}, and UCAST~\cite{ucast}.

The results of earlier baseline models~\cite{gatyscnnstyle, huangadain,liu2021adaattn,artflow} demonstrate that their portrayal of the intended texture style was limited, focusing only on color variations. And the content information is influenced by the content of the style image. 
Recent advances in traditional style transfer methods~\cite{mccnet_video,deng2021stytr2,Zhang:2022:CAST,CAP-VSTNet,ucast} show promise in capturing richer, more intricate textural nuances while better maintaining the structural integrity of the source content. Nevertheless, these state-of-the-art methodologies appear to struggle when the defining brushstrokes or stylistic elements within the reference artistic image are less prominently discernible, as exemplified by the inferior second-row outputs relative to other approaches.

In contrast, our method yields results with more intricate brushstrokes that closely resemble real art paintings. The major advantage of the diffusion approach lies in its capability to handle complex and non-stationary stylistic patterns. Unlike some general deep learning stylization methods that struggle with intricate or rapidly changing artistic styles, diffusion-based methods can effectively adapt to such variations. Diffusion methods are effective in preserving fine details and structural information and flexible in achieving desired artistic effects compared to general deep learning methods. These advantages make diffusion-based methods well-suited for producing high-quality, faithfully stylized images that capture the essence of various artistic styles.

\paragraph{Content modifications}

DiffStyler's dual diffusion is concerned with decoupling the representation of content and style. As a result, we can create more realistic, fine-grained structures in a variety of styles.
As shown in Fig.~\ref{fig:comparison}, our approach maintains better overall content compared to other approaches (e.g., the $1^{st}$ and $3^{rd}$ rows of Fig.~\ref{fig:comparison}).
In addition, to further verify the effectiveness of our method in content preservation, we also compare it with traditional style transfer methods.
As can be seen in Fig.~\ref{fig:comparison2}, DIffStyler can preserve both the detailed content and has a very harmonious overall effect. Balancing content and style preservation can be challenging for image style transfer. a significant advantage of DiffStyler is that there are no obvious artificial traces of content preservation of other traditional style transfer methods.
Finally, we experimentally found that the content modification is related to the chosen style, and is specified in Section~\ref{subsec:Discussion}.

\begin{figure*}
    \centering
    \includegraphics[width=\textwidth]{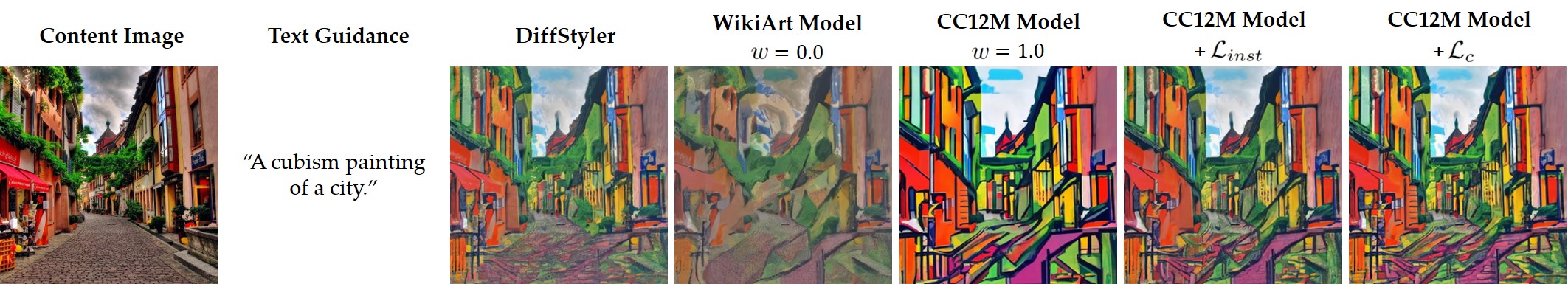}
    \caption{Ablation study of single and dual diffusion architecture.}
    \label{fig:ablation_dual}
\end{figure*}

\begin{figure*}
    \centering
    \includegraphics[width=0.8\linewidth]{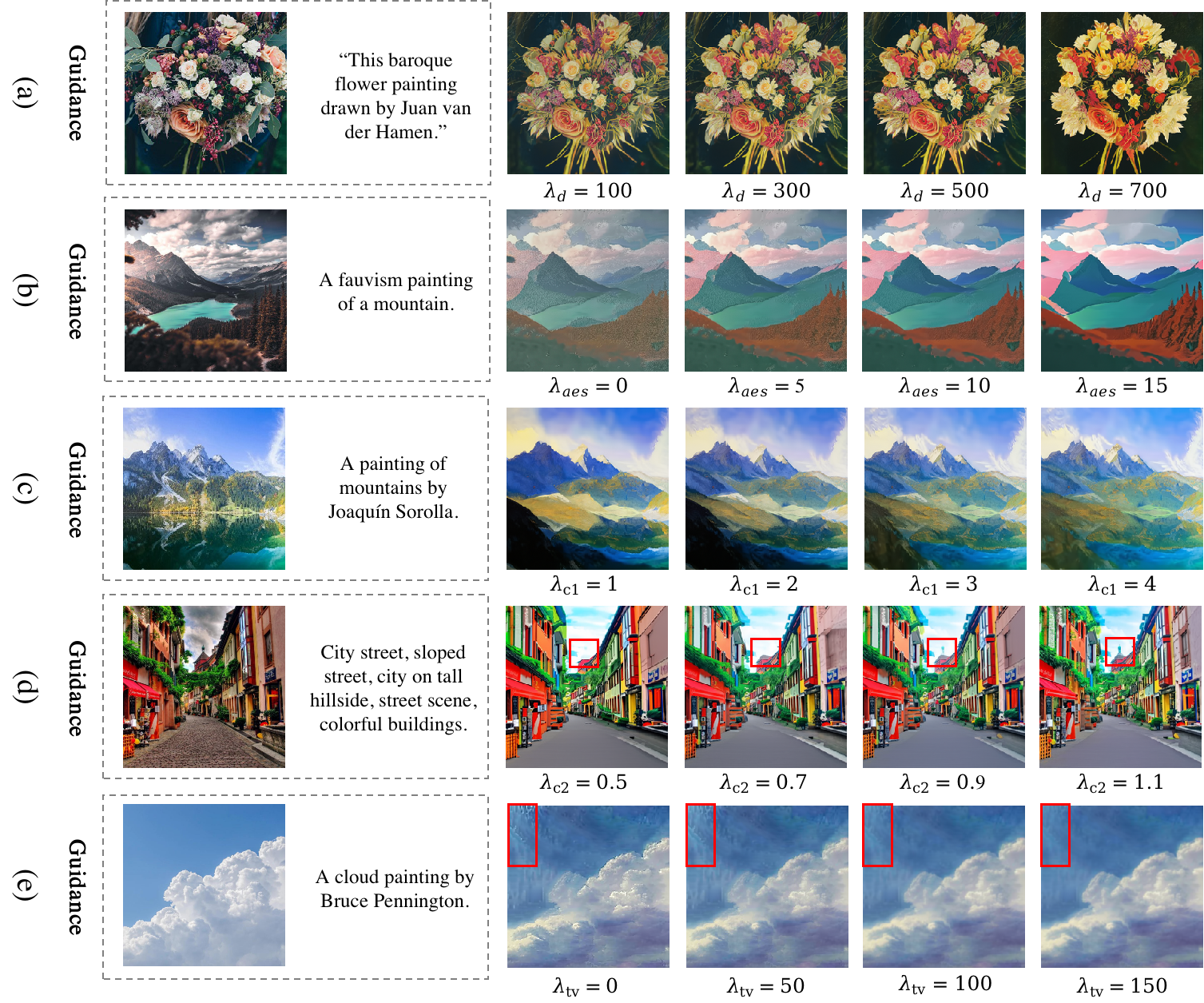}
    \caption{Results of ablation studies on loss functions. (a) Our full model; (b) without $\mathcal{L}_{\text {c}}$; (c) without $\mathcal{L}_{\text {c}_{patch}}$; (d) without $\lambda_{t v}$; and (e) without $\mathcal{L}_{a e s}$.}
    \label{fig:ablation_loss}
\end{figure*}

\subsection{Quantitative Evaluation}
For quantitative evaluation, we randomly selected $20$ content images and defined text prompts containing $25$ styles and artistic movement descriptions, generating $500$ stylized images for each method.
\paragraph{CLIP score}



To measure the correspondence between text prompts and stylized textures, we computed the CLIP score, which is the cosine similarity between the target texts derived from the CLIP encoder.
Since we employ the ViT-B/16 CLIP model throughout the inference phase, we compute the CLIP score using the ViT-B/32 CLIP~\cite{clip} model for fairness.

The results (the first row of Table~\ref{tab:quantity}) show that 
our method achieves the highest score of 0.2869. Stable diffusion \cite{latentdiffusion} follows closely with a score of 0.2773, marking the second-best result. This indicates that our approach demonstrates superior capability in expressing textual requirements through stylized image outputs.


\paragraph{LPIPS metric}
The learned perceptual image patch similarity (LPIPS)~\cite{LPIPS} metric is widely used to measure the difference between two images based on learned perceptual image patch similarity, which aligns more closely with human perception compared to traditional methods. The lower the LPIPS value, the higher the similarity between the images.

Analyzing the second row of Table~\ref{tab:quantity}, it can be observed that both the CLIPstyler (fast) and the DiffStyler methods outperform the other approaches in terms of content retention, as indicated by their lower LPIPS values. The proposed methods effectively preserve the content of the input images while applying the desired style. This performance is desirable, as it ensures that the stylized images maintain the key features and characteristics of the original content.

\paragraph{Content loss}
To perform a quantitative evaluation of our approach, we leverage content loss~\cite{content_loss} alongside establishing the state-of-the-art stylization techniques employed for test ratings. Table~\ref{tab:quantity_tra} presents the reported scores for both our proposed DiffStyler method and the baseline models. Notably, our approach showcases several advantages in terms of content preservation, even when compared to traditional stylization methods. The employed learnable noise methods successfully retain the essential content of the input images while effectively applying the desired style.


\begin{table*}[th]
  \centering
  \caption{User study results.
  The best results are highlighted in \textbf{bold} while the second best results are marked with an \underline{underline}.}
    \begin{tabular}{l|ccccccc}
    \toprule
    \multicolumn{1}{c}{ } & Ours  & LDAST~\cite{fu2022ldast}       & CLIPstyler (opti)~\cite{CLIPstyler} & CLIPstyler (fast)~\cite{CLIPstyler} & DiffusionCLIP~\cite{DiffusionCLIP} &  Stable Diffusion~\cite{latentdiffusion} & MGAD~\cite{Huang2022MGAD} \\
    \midrule
    Quality     & \textbf{4.41 } & 3.77        & 3.75        & 4.08        & 3.82        & \underline{4.13}        & 4.10  \\
    \midrule
    Content     & \underline{3.97}        & 3.94        & 3.48        & \textbf{4.06 } & 3.65        & 3.83        & 3.73  \\
    \midrule
    Consistency & \textbf{4.39 } & 3.73        & 3.84        & 3.95        & 3.71        & \underline{4.12}        & 4.08  \\
    \bottomrule
    \end{tabular}%
  \label{tab:user_study}%
\end{table*}%

\begin{figure*}
    \centering
    \includegraphics[width=\textwidth]{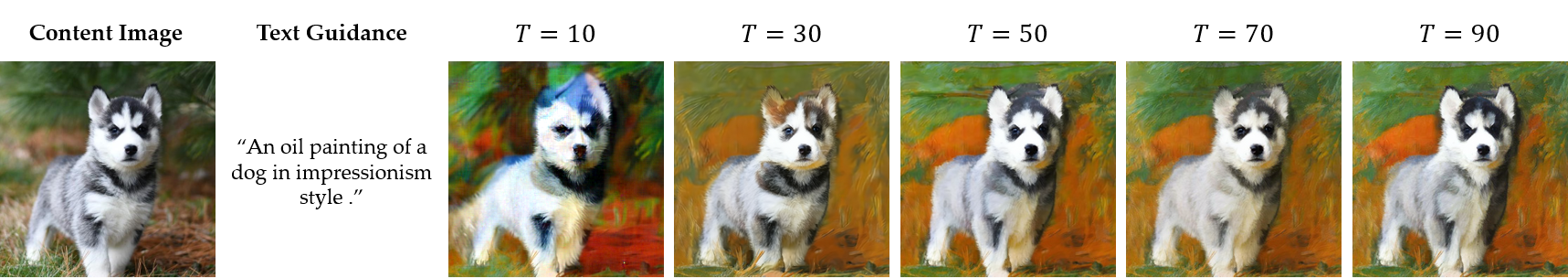}
    \caption{Results of different diffusion inference steps.}
    \label{fig:steps}
\end{figure*}

\paragraph{User study}
Since the evaluation of artistic aspects is relatively subjective, we conducted a user perception assessment to further compare our methods.
We compared DiffStyler with several state-of-the-art text-guided style transfer methods (i.e., including LDAST~\cite{fu2022ldast}, CLIPstyler~\cite{CLIPstyler}, DiffusionCLIP~\cite{DiffusionCLIP}, and Stable diffusion~\cite{latentdiffusion}) and multimodal driven image generation method MGAD~\cite{Huang2022MGAD}.
All baselines we used to perform inference were publicly available implementations with default configurations. For each participant, $28$ input-output pairs were randomly selected. Participants were requested to rate the following metrics on a $1-5$ scale based on the stylized results: (1) quality of stylized results, (2) content preservation, and (3) consistency of stylized results with text prompts. Finally, we collected $6720$ scored results from $80$ participants.
Table~\ref{tab:user_study} indicates that DiffStyler scores are high for all three evaluation metrics, which indicates that our findings excel at the level of text-guided stylization.

\begin{table}[htbp]
  \centering
  \caption{Ablation study on content preservation.}
    \begin{tabular}{c|c|c|c}
    \toprule
    Learnable noise & $\mathcal{L}_{\text {c}}$        & $\mathcal{L}_{\text {c}_{patch}}  $  & \multicolumn{1}{l}{LPIPS↓} \\
    \midrule
    \midrule
    \XSolidBrush & \Checkmark & \Checkmark & 0.674  \\
    \midrule
    \Checkmark& \XSolidBrush & \Checkmark & 0.593   \\
    \midrule
    \Checkmark & \Checkmark & \XSolidBrush & 0.521  \\
    \midrule
    \Checkmark & \Checkmark & \Checkmark & 0.505  \\
    \bottomrule
    \end{tabular}%
  \label{tab:ablation}%
\end{table}%

\subsection{Ablation Study}
\label{subsec:ablation}
\paragraph{Dual diffusion guidance scales}
The guidance scale $w$ of the diffusion model determines how strongly each of the two-path diffusion guides the final stylization outputs.
As shown in Fig.~\ref{fig:ablation_dual}, when the CC12M model is not utilized and only the WikiArt model is employed (as shown in column 4), the stylization effect closely aligns with the text guidance. However, this approach results in significant content deformation while expressing the desired style. On the other hand, when solely relying on the CC12M model without the WikiArt model (as shown in column 5), the generated image retains clearer content and exhibits a more pronounced style expression. Nevertheless, the generated image, when solely relying on the CC12M model without the WikiArt model, deviates from the style explicitly specified in the textual guidelines. Additionally, it displays issues such as over-saturated colors, unnatural textures, and noticeable artifacts. Furthermore, columns 6 and 7 of Fig.~\ref{fig:ablation_dual} highlight that the impact of the loss on image generation outcomes cannot be equated with the impact of the U-Net model alone. These results demonstrate the importance of employing the dual diffusion model architecture.

\paragraph{Loss terms}
We ablate the different loss terms in our objective by qualitatively comparing our results when training with our full objective (Equation~\ref{Eq_loss}) and with a specific loss removed. The results are shown in Fig.~\ref{fig:ablation_loss}.
We set up four control groups for ablation experiments to illustrate the effectiveness of loss. $\lambda_d$ is used as a control variable for $\mathcal{L}_{\text {inst }}$, and the generated results are more consistent with the text as $\lambda_d$ increases. $\lambda_{\text {c1}}$ is the control variable of $\mathcal{L}_{\text {c}}$, which makes the overall features of the generated image more consistent with the content map. $\lambda_{\text {c2}}$ is the control variable of $\mathcal{L}_{\text {c}_{patch}}$ for better performance of the detailed features. 
As shown in Table~\ref{tab:ablation}, without $\mathcal{L}_{\text {c}}$ and $\mathcal{L}_{\text {c}_{patch}}$, the LPIPS increased by $0.088$ and $0.016$.
$\lambda_{a e s}$ makes the results more consistent with human aesthetic preferences. The $\mathcal{L}_{t v}$, which is common in image generation work, can remove artifacts and improve smoothing as $\lambda_{t v}$ increases.
Thus, when we use all the proposed loss functions, we can obtain better-stylized results in the perceptual domain.



\begin{figure*}[btp]
    \centering
    \includegraphics[width=\linewidth]{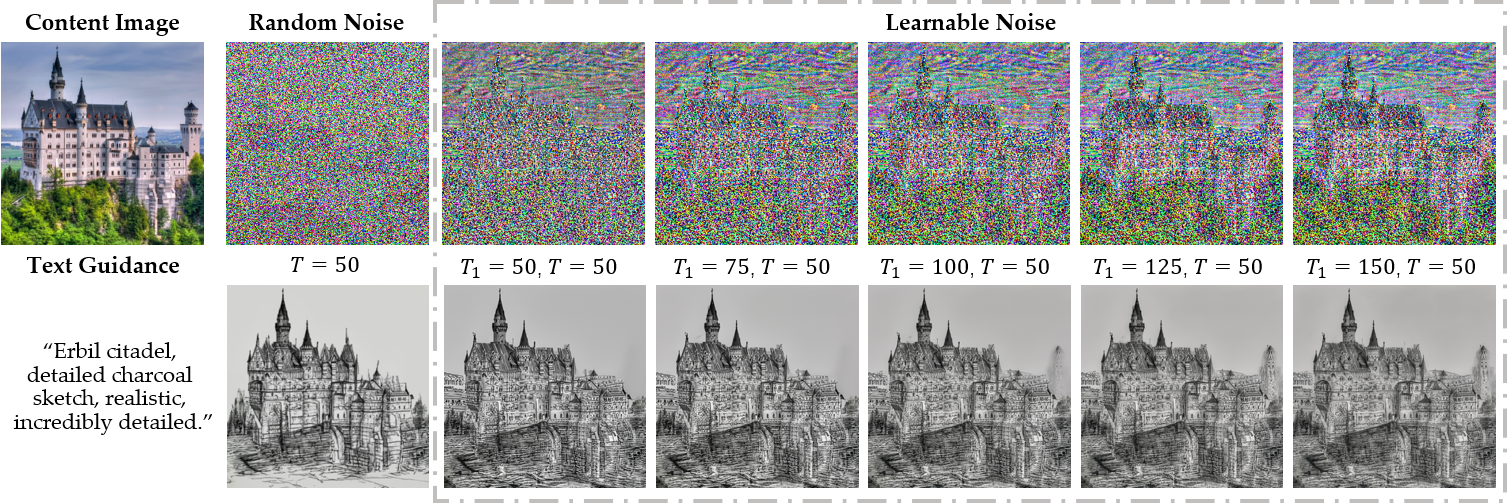}
    \caption{The introduction of noise induces alterations in image features, thereby influencing the reconstruction outcomes of the image content. The visualization graph provides a comparative analysis of the impact of random noise and learnable noise on the resulting content of the produced images.}
    \label{fig:learnablenoise}
\end{figure*}

\paragraph{Diffusion steps}
The number of steps determines the quality of the generation and the length of the generation time. We found experimentally (see Fig.~\ref{fig:steps}) that the number of inference steps $T$ is set to $50$, and we can obtain good results. When the number of inference steps is larger than $50$, the result changes are no longer significant and converge to the upper mass limit.
To maintain the structure of the input content, we find it beneficial when the number of free diffusion steps $T_1$ of learnable noise is more than the number of reverse sampling steps $T$. Ablation study results (Fig.~\ref{fig:learnablenoise}) show that at $T=50$. 
As shown in Table~\ref{tab:ablation}, LPIPS increased by $0.169$ in the absence of learnable noise.
The content of the stylized results is better maintained as $T_1$ increases gradually, and the image quality no longer changes significantly after a certain number of diffusion steps.  
To ensure the quality of the results and the efficiency of the generation process, the number of free diffusion steps $T_1$ used in the experiments of this work is $150$ and the number of inference process steps $T$ is $50$.

\begin{figure*}
    \centering
    \includegraphics[width=\textwidth]{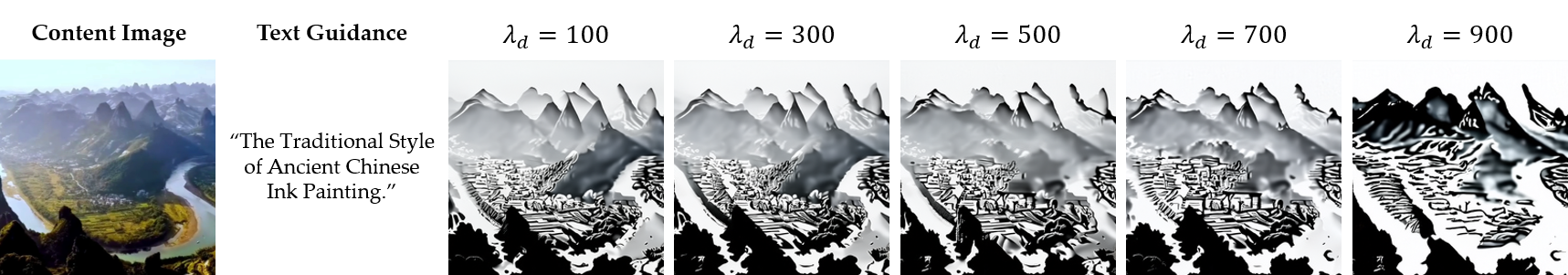}
    \caption{Effect of the CLIP guidance scale $\lambda_d$ on the generation results.}
    \label{fig:clip_guidance}
\end{figure*}

\paragraph{Effect of the CLIP Guidance Scale}
CLIP guidance scale $\lambda_d$ determines how consistently the results match the text prompts. We compared the generated results by taking different $\lambda_d$ values to verify the impact of text guidance. As shown in Fig.~\ref{fig:clip_guidance}, the generated results are more consistent with the input content image when using a smaller $\lambda_d$. As $\lambda_d$ is gradually increased, the generated results are brought closer to the semantic content of the prompt and the user's requirements. However, when the value of $\lambda_d$ is too large, the discrepancy between the stylized result and the content of the input image increases, which will cause the stylization to fail. Therefore, with an appropriate CLIP guidance scale value, the result can be made to match the text prompt content while keeping the content of the input image.

\begin{figure}[th]
    \centering
    \includegraphics[width=0.5\textwidth]{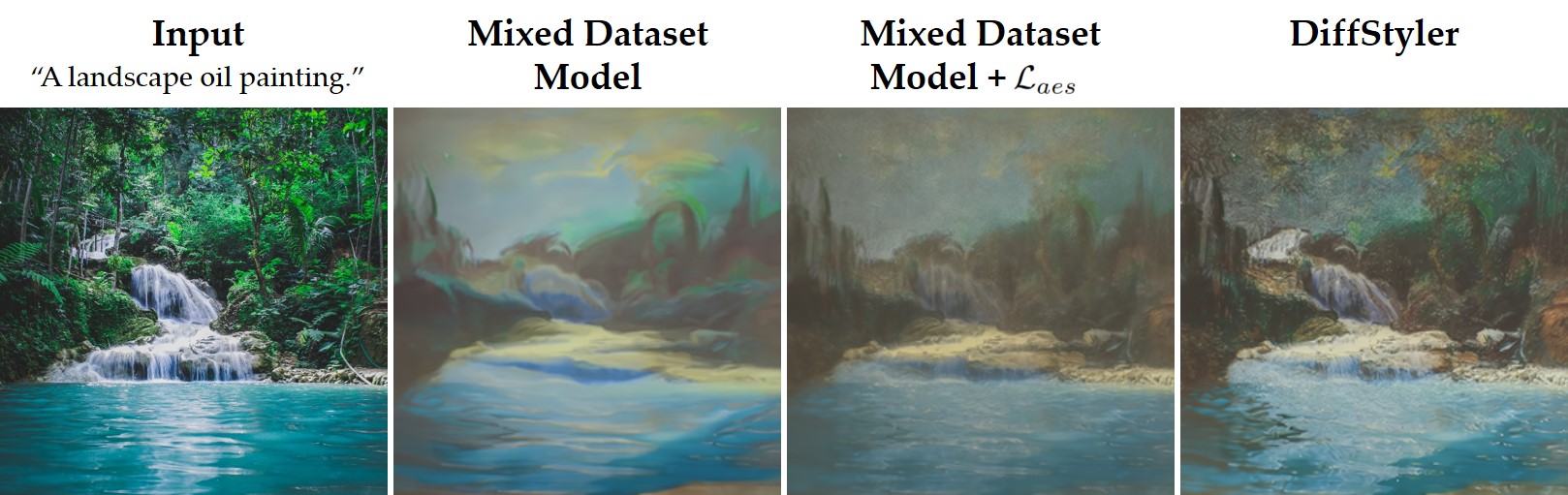}
    \caption{Ablation study of the mixed dataset results.}
    \label{fig:mixdataset}
\end{figure}
\paragraph{The mixed dataset results}
As shown in Fig.~\ref{fig:mixdataset}, we conducted ablation experiments on a mixed dataset. The inclusion of both the CC12M dataset, primarily composed of nature images, and the WikiArt dataset, consisting of art images, introduced a substantial scale disparity between the two datasets. Consequently, combining these datasets resulted in an imbalance in the expression of artistic features, wherein the art features were inadequately represented, and the potential of the art image dataset remained underutilized. This challenge was further exacerbated by the limited availability of online art images, hindering the comprehensive representation of diverse art features, as demonstrated by the aforementioned findings in our research. To tackle this issue, we propose a two-tier architectural model that aims to overcome the aforementioned limitations and improve the incorporation of art features in the stylization process.

\subsection{Discussion}
\label{subsec:Discussion}

\begin{figure*}
    \centering
    \includegraphics[width=0.78\linewidth]{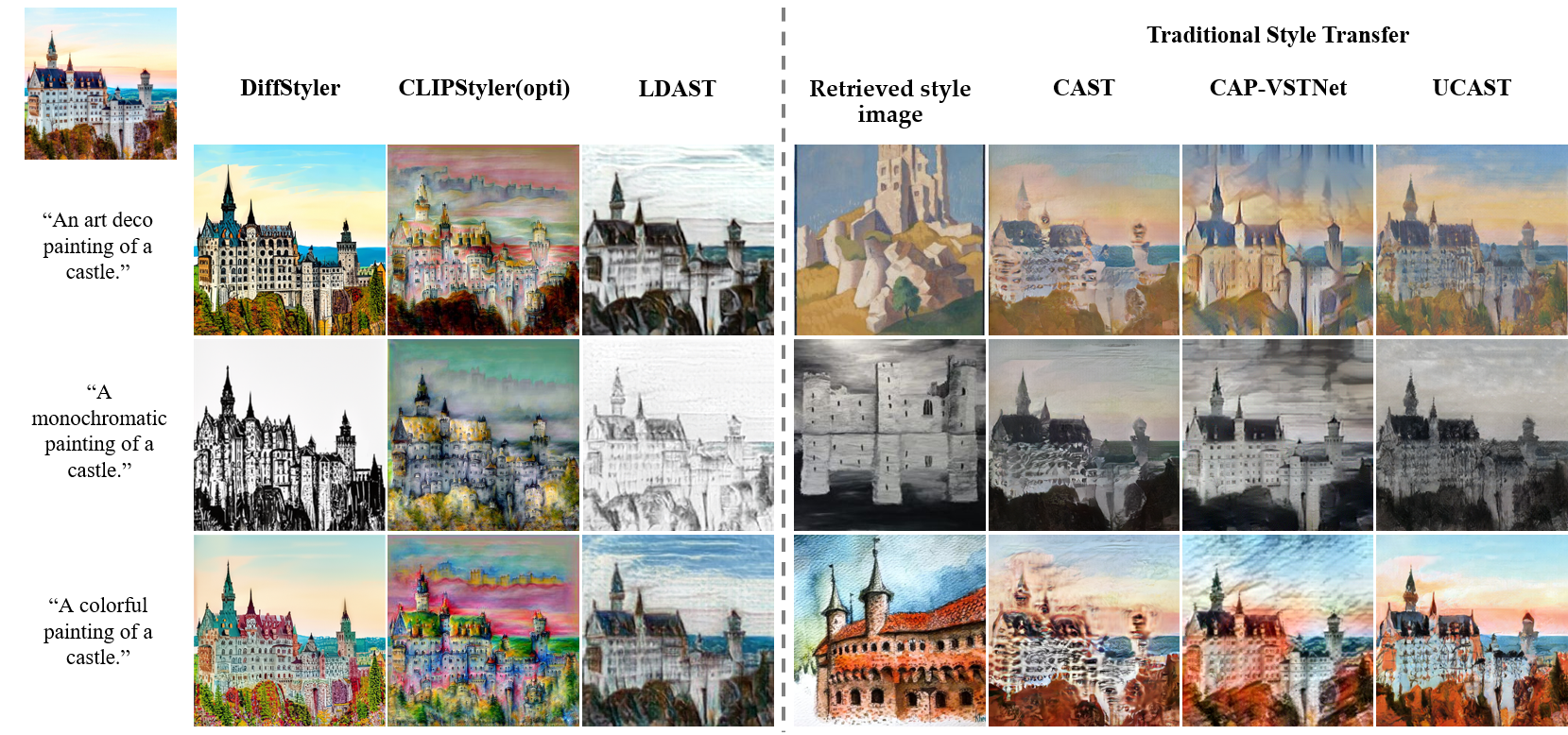}
    \caption{Comparison experiments for discussion.}
    \label{fig:discussion}
    \vspace{-0.5cm} 
\end{figure*}

Further experiments and comparisons revealed that the modification of the content is related to the chosen style. As Fig.~\ref{fig:discussion} shows, the content of the input image in the traditional styling task remains influenced by the style image, and the content remains influenced by the input text in the text-guided styling task. Compared with the SOTA image/text-guided style transfer approaches~\cite{CLIPstyler,fu2022ldast,Zhang:2022:CAST,CAP-VSTNet,ucast}, the dual diffusion pathway in DiffStyler focuses on the decoupling of content and style representation. As a result, we can generate more natural and fine-grained structures in different styles that are significantly better than other styles (e.g., castle windows and outlines). For a long time, balancing content and style preservation has been challenging for image style transfer. Nevertheless, with text input, the user can further improve the results by changing the text prompt.

\subsection{Failure cases}
\label{subsec:failurecases}
\begin{figure}[h]
    \centering
    \includegraphics[width=0.33\textwidth]{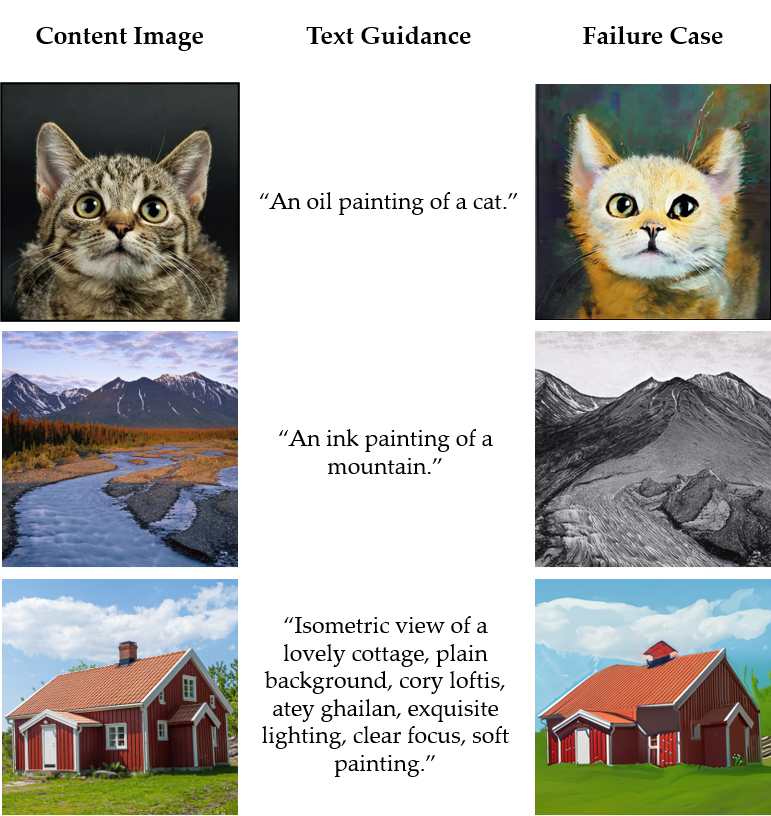}
    \caption{Failure cases of DiffStyler.}
    \vspace{-0.4cm} 
    \label{fig:failcase}
\end{figure}

While our approach demonstrates high-quality results for text-guided image stylization, it is important to acknowledge its limitations. As illustrated in Fig.~\ref{fig:failcase}, we present several instances where our model fails to produce satisfactory results. In the first row, there is a partial disappearance of the cat's fur; in the second row, the distinction between ``river'', ``land'' and ``mountain'' becomes less apparent; and in the third row, the windows of the house vanish. We have observed that the generated results primarily emphasize the elements mentioned in the text, potentially causing other elements to be overlooked and subsequently disappear.

In conclusion, these failure cases highlight the potential drawback of our approach, which relies on textual guidance. This reliance may lead to a limited focus on elements that are not explicitly mentioned or implicitly referenced. Consequently, important visual details or contextual relationships may be omitted, distorted, or inadequately represented. To address these limitations, it is crucial to enhance the model's capacity to capture and incorporate relevant contextual information beyond explicit textual guidance. This improvement would enable a more comprehensive and faithful stylization process.
\section{Conclusions and Future Work}
In this work, we propose a novel controllable dual diffusion framework, namely \shortname, for text-driven image stylization.
Our method contains a dual diffusion processing architecture to achieve a trade-off between abstraction and realism of the stylization results.
To eliminate the destructive effect of noise on the content of the input image, we use the results of the learnable diffusion process based on the content image to replace the random noise.
As the novel baseline for single text-driven style transfer using a diffusion model, the \shortname model has a highly distinguished ability to represent artistic style and provides new insights into the challenging problem of style transfer.
At present, the sample-time speed of our method is not as fast as some GAN-based approaches.
Performing diffusion processes in latent space to speed up the computation will be a possible future research direction.
Although we present high-quality results under a single text condition, our method is not without limitations. In particular, the input content may not be fully preserved in the results (see Section~\ref{subsec:failurecases}).
We plan to investigate more fine-grained control over stylization results in the future.

In addition, DiffStyler has not been able to implement video styling at this time, so we consider it very promising to extend the approach further into the video data domain.
We envision that possible future implementations are as follows: one approach is that noise can be used as crucial information for generating each frame in a video to achieve spatio-temporal consistency in video stylization. Another approach is that more realistic and high-quality video reconstruction can be achieved by introducing noise constraints in the video reconstruction process.
The potential for utilizing our approach to enhance the visual aesthetics and artistic expression of moving images is an exciting direction for future exploration.

\bibliographystyle{IEEEtran}
\bibliography{references}
\begin{IEEEbiography}[{\includegraphics[width=1in,height=1.25in,clip,keepaspectratio]{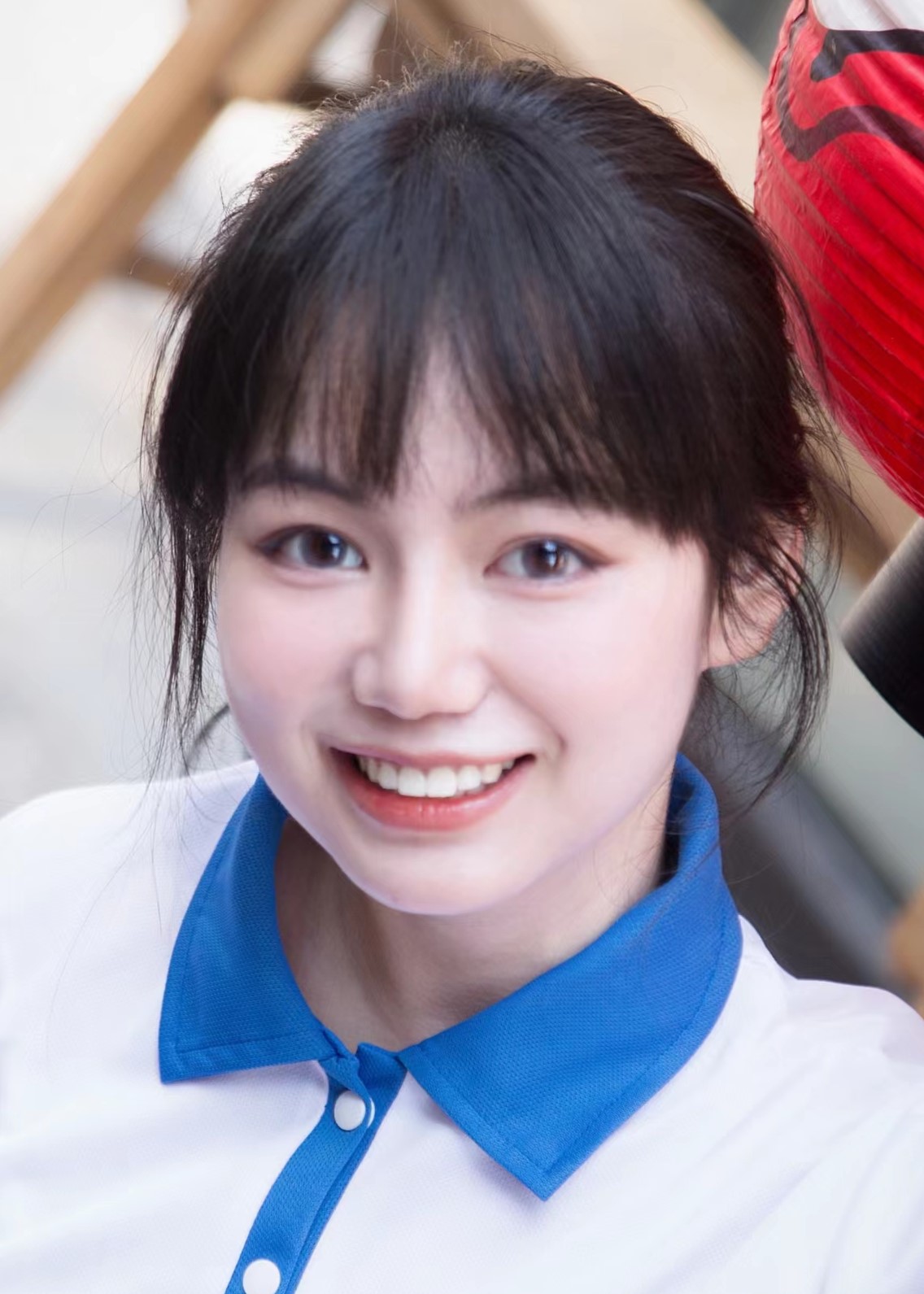}}]{Nisha Huang} received her BSc in Aerospace Information Engineering from Beihang University in 2021. She is now is pursuing the
M.E. degree at the State Key Laboratory of Multimodal Artificial Intelligence Systems (MAIS), Institute of Automation, Chinese Academy of Sciences, and the School of Artificial Intelligence at the University of Chinese Academy of Sciences. Her research interests include multimedia analysis, computer vision, and machine learning.
\end{IEEEbiography}
\begin{IEEEbiography}[{\includegraphics[width=1in,height=1.25in,clip,keepaspectratio]{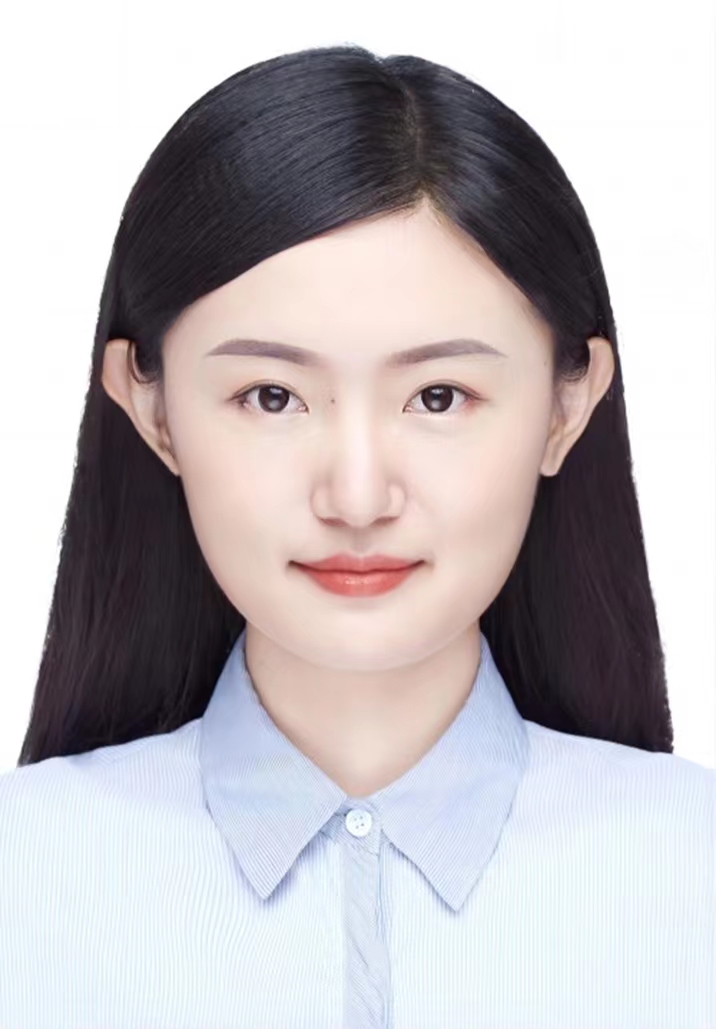}}]{Yuxin Zhang} received B.Sc. degree in Automation from Tsinghua University, Beijing, China, in 2020. She is now a Ph.D. candidate of the State Key Laboratory of Multimodal Artificial Intelligence Systems (MAIS), Institute of Automation, Chinese Academy of Sciences, and the School of Artificial Intelligence at the University of Chinese Academy of Sciences. Her research interests include computer vision, computer graphics, and machine learning.
\end{IEEEbiography}
\begin{IEEEbiography}[{\includegraphics[width=1in,height=1.24in,clip,keepaspectratio]{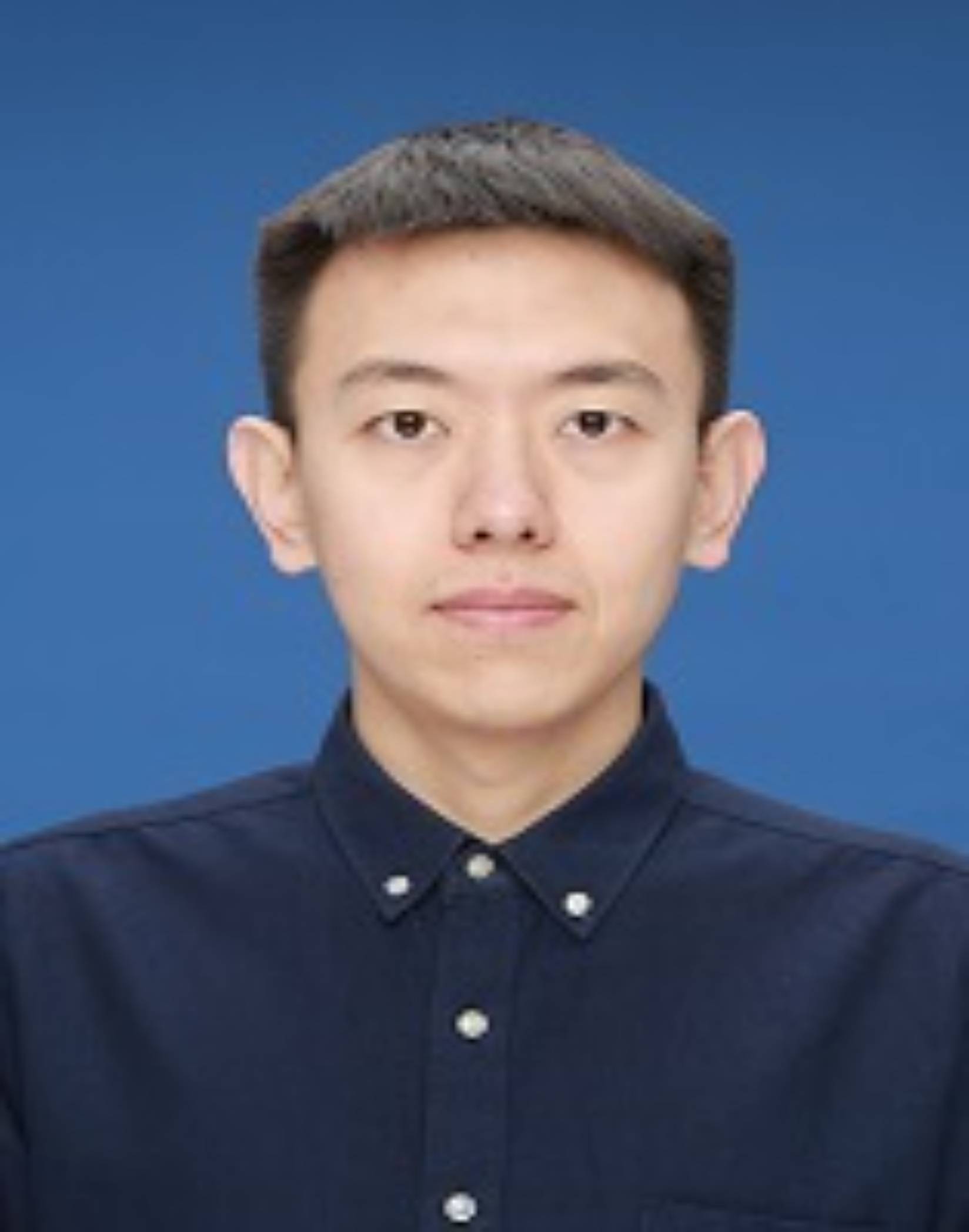}}]{Fan Tang} received the B.Sc. degree in computer science from North China Electric Power University, Beijing, China, in 2013, and the Ph.D. degree from Institute of Automation, Chinese Academy of Sciences, Beijing, in 2019. He is an Assistant Professor with the Institute of Computing Technology, Chinese Academy of Sciences. His research interests include computer graphics, computer vision, and machine learning.
\end{IEEEbiography}
\begin{IEEEbiography}[{\includegraphics[width=1in,height=1.25in,clip,keepaspectratio]{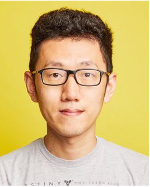}}]{Chongyang Ma}
received B.S. degree from the Fundamental Science Class (Mathematics and Physics) of Tsinghua University in 2007 and Ph.D. degree in Computer Science from the Institute for Advanced Study of Tsinghua University in 2012. He is currently a Research Lead with Kuaishou Technology, Beijing. His research interests include computer graphics and computer vision.
\end{IEEEbiography}
\begin{IEEEbiography}[{\includegraphics[width=1in,height=1.25in,clip,keepaspectratio]{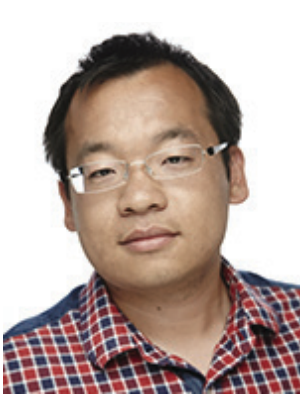}}]{Haibin Huang} received his BSc and MSc degrees in Mathematics in 2009 and 2011 respectively from Zhejiang University. He obtained his Ph.D. in Computer Science from UMass Amherst in 2017. He is currently a Senior Staff Research Scientist at Kuaishou Technology.
\end{IEEEbiography}
\begin{IEEEbiography}[{\includegraphics[width=1in,height=1.25in,clip,keepaspectratio]{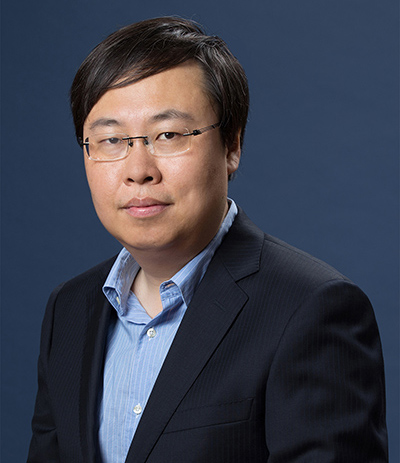}}]{Weiming Dong}
(Member, IEEE) is a Professor at the State Key Laboratory of Multimodal Artificial Intelligence Systems (MAIS), Institute of Automation, Chinese Academy of Sciences. He received his BSc and MSc degrees in 2001 and 2004, both from Tsinghua University, China. He received his PhD in Computer Science from the University of Lorraine, France, in 2007. His research interests include image synthesis, image recognition, and computational creativity.
\end{IEEEbiography}
\begin{IEEEbiography}[{\includegraphics[width=1in,height=1.25in,clip,keepaspectratio]{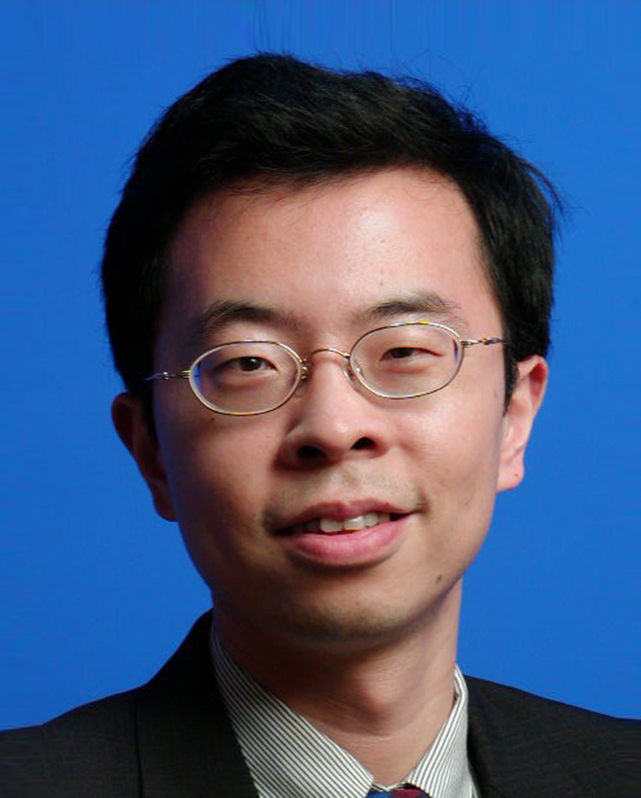}}]{Changsheng Xu}
(Fellow, IEEE) is a Professor at the State Key Laboratory of Multimodal Artificial Intelligence Systems (MAIS), Institute of Automation, Chinese Academy of Sciences.
Dr. Xu received the Best Associate Editor Award of ACM Transactions on Multimedia Computing, Communications and Applications in 2012 and the Best Editorial Member Award of ACM/Springer Multimedia Systems Journal in 2008. He has served as an Associate Editor, a Guest Editor, a General Chair, a Program Chair, an Area/Track Chair, a Special Session Organizer, a Session Chair, and a Transactions on Professional Communication (TPC) Member for over 20 IEEE and ACM prestigious multimedia journals, conferences, and workshops. He is the International Association for Pattern Recognition (IAPR) Fellow and the ACM Distinguished Scientist.
\end{IEEEbiography}
\end{document}